\documentclass[10pt,twoside]{article}       

\usepackage[dvipdfmx]{graphicx} 
\usepackage{bmpsize}

\usepackage{times}
\usepackage{url}
\usepackage{latexsym}
\usepackage{amssymb}
\usepackage{amsmath}
\usepackage{amsfonts}
\usepackage{paralist}  
\usepackage{verbatim}
\usepackage{multirow}
\usepackage{color}
\usepackage{comment}
\usepackage{rotating}
\usepackage{listings}
\usepackage{footnote}
\usepackage{arydshln}
\usepackage{natbib} 

\newcommand{\todo}[1]{\textcolor{red}{\noindent$\rightarrow$ TODO: #1}}

\newcommand{\commea}[1]{\textcolor{blue}{\noindent$\rightarrow$ COMMENT EA: #1}}

\graphicspath{{./figures/}}
\renewcommand{\S}[0]{Section~}

\begin{document}

\title{Evaluating the word-expert approach for Named-Entity Disambiguation}

\author{Angel~X. Chang$^\dagger$ \and Valentin~I. Spitkovsky$^\dagger$ \and Christopher~D. Manning$^\dagger$ \and Eneko Agirre$^\ddagger$ \and \\
           $^\dagger${\small Computer Science Department, Stanford University, Stanford, CA, USA}\\
           $^\ddagger${\small IXA Group, University of the Basque Country, Donostia, Basque Country}\\
           {\small \{angelx,vals,manning\}@cs.stanford.edu, e.agirre@ehu.eus}}


\date{}

\maketitle

\begin{abstract}\noindent
  Named Entity Disambiguation (NED) is the task of linking a
  named-entity mention to an instance in a knowledge-base, typically
  Wikipedia.  This task is closely related to word-sense
  disambiguation~(WSD), where the supervised word-expert approach has
  prevailed.  In this work we present, for the first time, the results
  of the word-expert approach to NED, where one classifier is built
  for each target entity mention string. The resources necessary to
  build the system, a dictionary and a set of training instances, have
  been automatically derived from Wikipedia. 
  We provide empirical
  evidence of the value of this approach, as well as a study of the
  differences between WSD and NED, including ambiguity and synonymy
  statistics.



\end{abstract}

\section{Introduction}
\label{sec:introduction}

Construction of formal representations from snippets of free-form text is
a long-sought-after goal in natural language processing~(NLP).
Grounding written language with respect to background knowledge
about real-life entities and world events is important 
for building such representations.  It also has many applications in
its own right: text mining, information retrieval and the semantic web~\citep{WeikumT10}.
Wikipedia and related  repositories of structured data (e.g.
WikiData or DBpedia) already provide extensive inventories of named entities, including
people, organizations and geo-political entities.

An individual named-entity string may refer to multiple entities and
the process of resolving the appropriate meaning in context is called
entity linking~(EL) or named entity disambiguation~(NED).  The former
terminology~(EL) stresses the importance of linking a mention to an
actual instance in the given knowledge-base~\citep{McNameeAndDang09}.
We prefer the latter term~(NED), which focuses on the potential
ambiguity among several possible instances. It highlights the
connection to the closely related problem of word-sense disambiguation
(WSD), and was used in some of the earliest
works~\citep{DBLP:conf/eacl/BunescuP06,DBLP:conf/emnlp/Cucerzan07}.

\begin{figure}[t!]
\includegraphics[width=\textwidth]{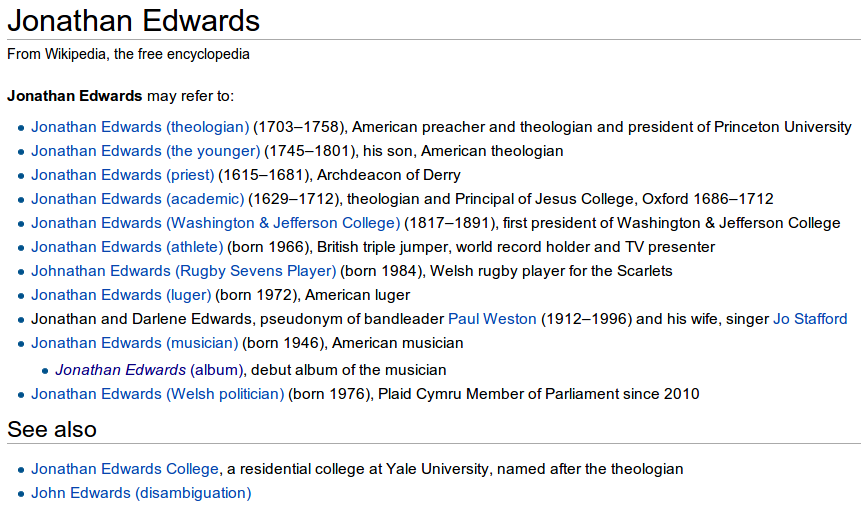}
\caption{Wikipedia disambiguation page for Jonathan Edwards.}\label{fig:jonathandis}
\end{figure}

\begin{figure}[th]
\includegraphics[width=\textwidth]{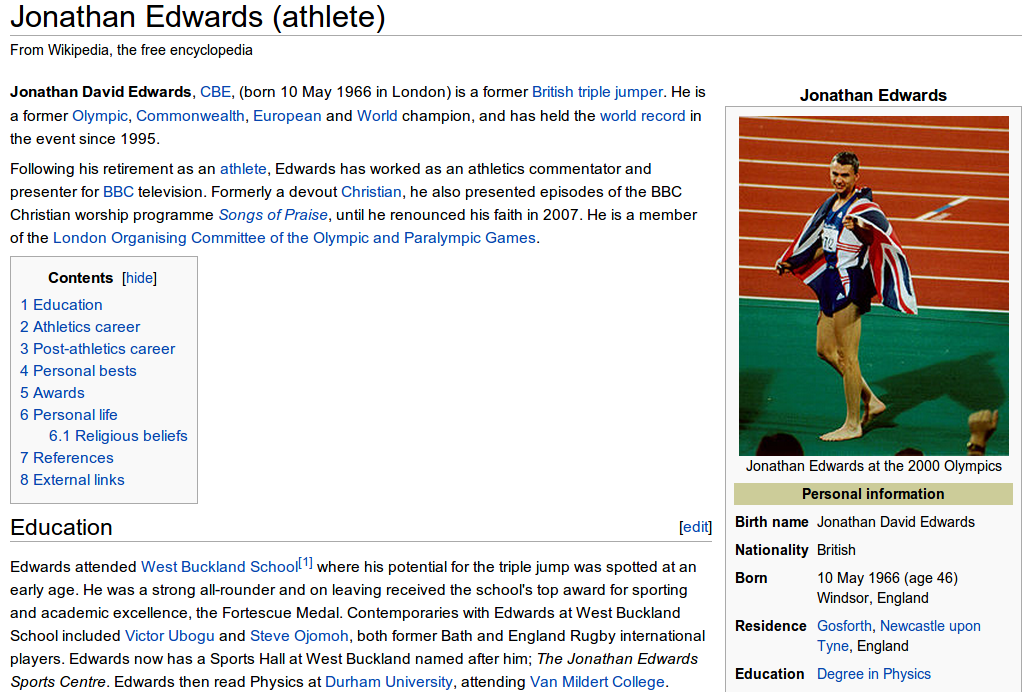}
\caption{Wikipedia article for Jonathan Edwards, the athlete.}\label{fig:jonathanathlete}
\end{figure}

As our first example of NED, consider the following sentence: \emph{``Champion
  triple jumper \textbf{Jonathan Edwards} has spoken of the impact
  losing his faith has had on his life.''}  Jonathan Edwards may
refer to several people --- Wikipedia lists more than
ten,\footnote{\url{http://en.wikipedia.org/wiki/Jonathan_Edwards_(disambiguation)}}
including the intended athlete\footnote{\url{http://en.wikipedia.org/wiki/Jonathan_Edwards_(athlete)}} ---
as well as a residential college at Yale and a music
record. Figure~\ref{fig:jonathandis} shows a disambiguation page for the string,
and Figure~\ref{fig:jonathanathlete} an excerpt from the athlete's Wikipedia article.

\begin{figure}[th]
\begin{center}
\begin{tabular}{c|p{9cm}}
\small{\em{Wikipedia article}} & \small{\em{Example sentence}} \\ 
\footnotesize{Triple\_Jump} & \scriptsize{The current male and female world record holders are \underline{Jonathan Edwards} of Great Britain, with a jump of 18.29 meters, and Inessa Kravets of Ukraine, with a jump of 15.50 m.}\\
\footnotesize{Ilfracombe} & \scriptsize{During the lead up to the 2012 Olympics a number of people carried the torch through Ilfracombe, amongst these was \underline{Jonathan Edwards}, the triple jump world record holder, who had the privilege to carry the torch past his former home.}\\
\footnotesize{Great\_North\_Museum} & \scriptsize{Athlete \underline{Jonathan Edwards} is the patron of the 'Be Part of It' campaign.} \\
\footnotesize{Christian\_Olsson} & \scriptsize{Olsson first became interested in triple jump after watching \underline{Jonathan Edwards} set the world record at the World Championships in his hometown Gothenburg.} \\
\footnotesize{Test\_the\_Nation} & \scriptsize{They were pitted against a group of 10 celebrities including EastEnders star Adam Woodyatt, pop music presenter Fearne Cotton, Olympic athlete \underline{Jonathan Edwards} and former Sunday Times editor and broadcaster Andrew Neil.}\\
\end{tabular}
\end{center}
\caption{Example sentences from Wikipedia whose anchor-texts point to \protect\url{Jonathan_Edwards_(athlete)}.}\label{fig:johnexamples1}
\end{figure}

\begin{figure}[th]
\begin{center}
\begin{tabular}{c|p{3.6cm}|c}
\small{\em{Wikipedia article}} & \small{\em{Example sentence}} & \small{\em{Target Wikipedia article}}\\ 
\footnotesize{Mission\_(Christian)} & \scriptsize{In North America, missionaries to the native Americans included \underline{Jonathan Edwards}, the well known preacher of the Great Awakening , who in his later years retired from the very public life of his early career.} & \footnotesize{Jonathan\_Edwards\_(theologian)} \\
\footnotesize{One\_Day\_Closer} & \scriptsize{One Day Closer is the ninth studio album (eleventh total album) released by singer songwriter \underline{Jonathan Edwards} and features many ballads and love songs.} & \footnotesize{Jonathan\_Edwards\_(musician)} \\
\footnotesize{Buildings\_of\_Jesus\_College,\_Oxford} & \scriptsize{\underline{Jonathan Edwards} (principal from 1686 to 1712) is reported to have spent \pounds 1,000 during his lifetime on the interior of the chapel.} & \footnotesize{Jonathan\_Edwards\_(academic)} \\
\end{tabular}
\end{center}
\caption{Example sentences from Wikipedia where the referent of ``Jonathan Edwards'' is not the athlete.}\label{fig:johnexamples2}
\end{figure}

\begin{figure}[th]
\begin{center}
\begin{tabular}{c|p{8cm}}
\small{\em{Wikipedia article}} & \small{\em{Example sentence}} \\ 
\footnotesize{Arthur\_Levitt} & \scriptsize{Levitt was appointed to his first five-year term as Chairman of the SEC by \underline{President Clinton} in July 1993 and reappointed in May 1998.} \\
\footnotesize{Asa\_Hutchinson} & \scriptsize{Hutchinson, who had at first decided to run for an open seat in the Arkansas House of Representatives from Sebastian County, defeated Ann Henry, a long-time friend of \underline{Bill} and Hillary Clinton}. \\
\footnotesize{William\_Jefferson\_Blythe,\_Jr.} & \scriptsize{Three months later, Virginia gave birth to their son, \underline{William Jefferson Bl}y\underline{the III}, the future President.}\\ 
\footnotesize{HMMT-164} & \scriptsize{In February 1996, HMM-164 was called upon to support the \underline{President of the United States} as he visited the flood-ravaged areas around Portland, Oregon.}\\
\footnotesize{Joan\_Jett\_Blakk} & \scriptsize{Smith also ran for president in 1996 with the slogan ``Lick \underline{Slick Willie} in '96!''} \\
\end{tabular}
\end{center}
\caption{Example sentences from Wikipedia that use different names to refer to the same entity (Bill Clinton).}\label{fig:billclintonexamples}
\end{figure}

In addition to disambiguation pages that list possible entities
to which a canonical string like ``Jonathan Edwards'' may refer, many
naturally-occurring entity
mentions in regular Wikipedia articles are also cross-referenced.
For instance, the first sentence of the entry for
Jonathan Edwards (the athlete) includes hyperlinks (shown in blue)
to ``CBE,'' ``British'' and ``triple jumper.'' The first link references an article on
the Order of the British
Empire.\footnote{\url{http://en.wikipedia.org/wiki/Order_of_the_British_Empire}}
\emph{Anchor-text} (words in blue) often exposes alternate
ways of referring to entities: e.g., a member of the
Order of the British Empire can be called ``CBE.'' Figure~\ref{fig:johnexamples1}
shows five sentences that link to Jonathan Edwards (the athlete), and Figure~\ref{fig:johnexamples2}
shows three sentences with hyperlinks to other people called Jonathan Edwards. Anchors present a
rich source of disambiguation information. Aggregating over all occurrences of hyper-text
``Jonathan Edwards,'' we can compute that, most of the
time, it refers to the theologian. Yet by analyzing the context of each named entity's occurrence,
we could conclude that a span like \emph{``\ldots triple jumper
Jonathan Edwards \ldots''} is more similar to mentions of the athlete.
These linked spans from Wikipedia can also be used to obtain alternative terms that refer to the same entity. 
Figure~\ref{fig:billclintonexamples}
shows a variety of ways in which Wikipedia refers to Bill Clinton.

As online encyclopedias grow in size, entities and ideas that are of interest to even
small communities of users may get their own Wikipedia pages with relevant descriptions.\footnote{\url{http://en.wikipedia.org/wiki/Wikipedia:Size_of_Wikipedia}}
Thanks to the hyper-linked nature of the web,
many online mentions have already been annotated with
pointers to corresponding articles, both within Wikipedia and from external sites.
As a result, large quantities of freely available information --- suitable
for supervised machine learning algorithms --- already exist,
obviating the need for costly manual annotations that are typically
associated with the training of traditional NLP systems in general and WSD in particular.

NED is a disambiguation task that is closely related to WSD, where the
goal is to disambiguate open-class words~(i.e., common nouns,
adjectives, verbs and adverbs) with respect to a sense inventory, such
as WordNet~\citep{Fellbaum:98}.  The extensive WSD literature
~\citep{AgirreAndEdmonds2006a,Navigli:2009:WSD:1459352.1459355} has
shown that building a supervised classifier for each target lemma --- the so-called word-expert approach ---
outperforms other techniques~\citep{zhong-ng:2010:Demos}.




In this work we propose an architecture for NED following the
word-expert approach, where we build a classifier for each
named-entity mention, with two main modules: (1) A candidate
generation module which, given a string, lists all potentially
relevant entities. This module is based on a static dictionary, which
also lists the popularity of each of the entities, and can thus serve
as a standalone context-independent disambiguation module. (2) A
context-sensitive supervised classifier that selects the entity which
is most suited for the context of the mention. The classifiers use the
kind of features routinely used in a WSD classifier. To our knowledge,
this is the first time a NED system following the word-expert approach
is reported, although early work on Wikification already hinted at its
usefulness~\citep{10.1109/MIS.2008.86}.

Our system is based on
two main resources: 
(i) a
dictionary listing all candidate entities for each surface string,
together with their
frequencies;\footnote{\url{http://www-nlp.stanford.edu/pubs/crosswikis-data.tar.bz2}}
and (ii) a set of training instances for each target mention.
The bulk of the information comes from Wikipedia but we
have further complemented the dictionary with web counts from a subset
of a 2011 Google crawl. The dictionary's release was documented in a
short conference paper~\citep{SPITKOVSKY12.266.L12-1109}, which we
extend with additional explanations and analyses here.


We present a detailed analysis of the performance of the
components and variations of our NED system, as applied to the entity
linking task of the NIST Text Analysis
Conference's~(TAC)\footnote{\url{http://www.nist.gov/tac/}}
knowledge-base population~(KBP)
track~\citep{ji-grishman:2011:ACL-HLT2011}.  The task focuses on
several target entity mentions, which makes it well suited for our
word-expert approach. In the future we would like to explore other
datasets which include all mentions in full documents \citep{hoffart-EtAl:2011:EMNLP}.
%
Our final results are quite strong despite the simplicity of the
techniques used, with the dictionary's raw frequencies already
performing extremely well.  We focus on the candidate generation and
disambiguation modules, leaving aside mention detection and the task
of NIL detection, where mentions which refer to entities not listed in
the knowledge-base have to be detected~\citep{DurrettKlein2014}.


In addition, we will study differences between the closely related
worlds of WSD and NED.  In WSD, an exhaustive dictionary is provided,
while in NED, one has to generate all candidate entities for a
target string --- a step that has been shown to be critical to
success~\citep{Hachey2012}.  In WSD very few occurrences
correspond to senses missing in the dictionary, but in NED this problem is
quite prevalent. We will also show that ambiguity is larger for
NED. On the bright side, there is a lot of potential training data
for NED, for instance, in the form of human-generated
anchor-texts. This article shows that an architecture based on WSD
methods can work well for NED, and that it is feasible to model
candidate generation with a static dictionary.  We will also compare
ambiguity, synonymy and inter-annotator agreement statistics of both
problems.

Note that the authors participated in the TAC-KBP Entity Linking
tracks with preliminary versions of the system reported in this
article. Those systems, alongside all participant
systems\footnote{There was no peer-review, and all papers were
  accepted.}, were reported in publicly available
proceedings \citep{agirre2009tackbp,agirre2010tackbp,agirre2011tackbp}.

The paper is organized as follows:  we first present
related work, followed by the architecture of
our system~(\S\ref{sec:architecture}).  The dictionary for candidate generation is presented
next~(\S\ref{sec:dictionary}), followed by methods that produce the training
data and build the supervised classifier~(\S\ref{sec:superv-disamb}).
\S\ref{sec:dataset} introduces the TAC-KBP dataset used for evaluation,
followed by the adaptation of our system~(\S\ref{sec:ned-system-tac}).
\S\ref{sec:development} explores
several alternatives to our approach and analyzes their performance on
development data, as well as the final results and discussion.
We then compare our system to related work,
draw conclusions and propose future directions.

\section{Related Work}
\label{sec:previous-work}

We will now review several NLP problems that are closely related to NED,
including wikification and WSD, as well as previous work on NED.  For clearer exposition, we group the previous work into three
sections: first, the earlier and more influential
contributions, followed by work on wikification, and finally 
 NED systems.  We briefly touch on state-of-the-art techniques
used in NED, specifically, handling of candidate generation (i.e., our dictionary),
and disambiguation.
A full comparison of our results
to those of the latest state-of-the-art systems can be found in \S\ref{sec:comparative-results}.

\subsection{Related Problems}
\label{sec:word-sense-disamb}

 

NED is related to several problems in NLP.  For instance, it
presupposes that mentions of named entities have already been
identified in text, building up from the named entity recognition~(NER)
task~\citep{Marsh98,TjongKimSang:2003:ICS:1119176.1119195}.  Each mention
may be further labeled with a broad semantic category --- such as names
of persons, organizations or locations ---
via named entity classification~(NEC). This last task is often
performed by using gazetteers to cover many known entities,
in addition to training a single supervised classifier that
outputs category types for input mentions given their specific contexts.
Overall, we view NED as a specific instantiation of the record linkage problem, in which the task is to find records referring to the
same entity across different data sources, such as data files,
databases, books or websites.

The term \emph{record linkage} was first used by
~\citet{citeulike:642364}, in reference to resolving person
names across official records held by a government. 
More recently, \citet{bagga-baldwin:1998:ACLCOLING} focused on cross-document
coreference of people by first identifying coreference chains within
documents and then comparing the found chains' contexts
across documents. 
Similar problems arise in citation
databases, where it is necessary to decide which mentions
of authors in bibliographic records refer to the same
person~\citep{bhattacharya:tkdd07}.
Other typical applications that lack a predefined inventory of entities
include resolving names in e-mails and web people search~(WePS).
In the WePS task, starting from a set of web-pages that
mention a name (e.g., John Doe), the goal is to decide how many John
Does there are, and who is mentioned
where~\citep{Artiles:2008:WPS:1367497.1367661,Artiles2009}.  
The e-mails task can be tackled by assuming that
each address corresponds to a distinct person and that
people's identifying information can be deduced from what
they write~\citep{elsayed:acl08}.


%
Three important aspects differentiate NED from record linkage and
other cross-document entity coreference tasks, such as the exercise
studied at ACE 2008\footnote{\url{http://www.itl.nist.gov/iad/mig//tests/ace/2008/doc/}}
and clustering of documents that mention the same entity~\citep{mann2003conll,gooi-allan:2004:HLTNAACL}.
These differences hinge on the existence of (1)~a
knowledge-base~(e.g., Wikipedia) that lists gold-standard
entities; (2)~rich textual information describing each entity~(i.e., its Wikipedia page);
and (3)~many explicitly disambiguated mentions of entities~(i.e., incoming hyper-links
furnished by Wikipedia's contributors and other, external web publishers).
A more closely related task is wikification~\citep{DBLP:conf/cikm/MihalceaC07,10.1109/MIS.2008.86,Milne:2008,Kulkarni:2009:CAW:1557019.1557073}, which involves first deciding which keywords or concepts are
relevant in a given text and then disambiguating them by linking to the
correct Wikipedia article.  Although the overall thrust of that task is
different, since wikification systems target
common nouns as well as named entities, 
its disambiguation techniques are relevant to NED~(see \S\ref{sec:wikification}).


\subsection{Word Sense Disambiguation}
\label{sec:word-sense-disamb-1}

In WSD, the task is to determine which
sense of an open class content word --- i.e., an adverb, verb, adjective or noun that isn't a named entity
--- applies to a particular occurrence of that
word~\citep{AgirreAndEdmonds2006a,Navigli:2009:WSD:1459352.1459355}.
Typically, the sense inventory is taken from a dictionary, such as WordNet~\citep{Fellbaum:98},
and fixed in advance.  Dictionaries are comprehensive, covering nearly
all uses of a word.  Consequently, WSD systems tend to return a sense for
every occurrence. For instance, the Senseval-3 \emph{lexical sample} dataset
\citep{chklovski-EtAl:2004:Senseval-3} contains 2,945 manually annotated
occurrences, 98.3\% of which have been assigned a dictionary sense:
only 1.7\% are problematic and have senses not found in the
dictionary. By contrast, in NED, a significant portion of mentions
are out-of-inventory (around 50\% in our dataset). 
However, in WSD target senses are often open to interpretation,
as reflected by low inter-annotator agreement, e.g., 72.5\% in
the Senseval-3 all-words task~\citep{snyder-palmer:2004:Senseval-3}.
We will show that NED poses a better-defined problem, with less dispute about what constitutes a correct target
entity, as indicated by significantly higher inter-annotator agreement.

WSD and NED differ also in two other key properties: (i) polysemy, the expected number of senses a word might have;
and (ii) synonymy, the expected number of different words that may be used to lexicalize a given concept.  These
statistics can be used to characterize the difficulty of an evaluation set.  We compared average
polysemy and synonymy values for WSD and NED, computed using gold
standards (by counting senses listed in dictionaries
versus those actually occurring in test data).
NED scored substantially higher on both metrics, relative to WSD~(see Section~\ref{sec:ambig-synonymy-dict}).  

The best-performing WSD systems are currently based on
supervised machine learning, judging by public evaluation
exercises~\citep{snyder-palmer:2004:Senseval-3,pradhan07}.
Typically, the problem is modeled using multi-class
classifiers~\citep{supchapter}, with one
classifier trained for each target word~(a.k.a. \emph{the word expert approach}). 
Training examples are represented by feature vectors~(see \S\ref{sec:superv-disamb}) and labeled
with gold senses.  At test time, inputs are processed and represented in the same way,
as vectors of features, with appropriate classifiers predicting intended senses.
Our proposed system architecture for NED follows this design~(see \S\ref{sec:architecture}).


WSD systems can perform well when training data are plentiful. On the Senseval-3 \emph{lexical sample}
task~\citep{chklovski-EtAl:2004:Senseval-3}, accuracies can reach as high as 72.9\% with fine-grained senses and 79.3\% with coarser-grained senses .  Both performance numbers are well above the most-frequent sense~(MFS)
baselines (55.2 and 64.5\%, respectively).
The test set used there comprised
57 target words, each backed by at least 100 manually-annotated examples.
But many words found in running text lack sufficient training instances, leading to lower
performance in evaluations when all words are considered.
The best accuracy reported for the Senseval-3 \emph{all words} task, testing
all open-class words occurring in three texts (editorial, news and fiction),
was 65.1\%, only slightly higher than the MFS baseline's 62.4\%~\citep{snyder-palmer:2004:Senseval-3}.
Unfortunately, it is expensive to produce thorough hand-tagged training data for WSD and people do not typically annotate their words with the exact sense they meant.

In contrast with WSD, where training data is scarce,
the upside for NED is that labeled instances for many mentions are
already available in large numbers, annotated by Wikipedia volunteers,
since the contexts and targets of hyper-links could be used as supervision.
In other respects, we find many similarities between WSD and NED:  For each
mention that could be used to refer to a named entity,
the knowledge base may list all possible targets, as the
dictionary does for WSD. E.g., for a string like ``John Edwards''
a disambiguation page may list all relevant entities~(see Figure~\ref{fig:jonathandis}).
Naturally, there are also differences: A big downside
for NED is that the inventory of
meanings is not explicit and, for most strings, will be badly incomplete,
unlike dictionaries used in WSD.

In Sections \ref{sec:dataset} and \ref{sec:development} we will explore
the relation between WSD and NED further, showing that %
(1)~ambiguity, synonymy and incidence of dictionary misses are all
higher for NED than for WSD;
(2)~the NED task appears better-defined, as signaled by
higher inter-annotator agreement than in WSD; %
(3)~the skew of frequencies is more extreme for NED, with MFS
consequently presenting an even stronger baseline there than in WSD; %
(4)~the high number of training instances available to NED enables better
supervised training, allowing NED systems to follow the same
architecture as WSD systems, using analogous
preprocessing, feature sets, and classifiers; %
and (5)~the high ambiguity of mentions encountered by NED makes a
typical word expert approach more computationally expensive,
but still feasible.  Lastly, (6) we will discuss the
feasibility of constructing a comprehensive dictionary for NED.

\subsection{Early Work on NED}

The earliest work on NED using Wikipedia is by
\citet{DBLP:conf/eacl/BunescuP06}, who used
article titles, redirects and disambiguation pages
to generate candidate entities.  Similarity between
a mention's context and article text provided the
rankings, according to tf-idf and cosine similarity.
Each article's term vector was further enriched
using words from other articles in the same category.
Disambiguation of mentions was \emph{local}, i.e.,
performed separately for each one.

\citet{DBLP:conf/emnlp/Cucerzan07} followed an overall similar design, but using
context vectors that consisted of key words and short phrases extracted
from Wikipedia.  He disambiguated all named entities in text simultaneously,
adding a \emph{global} constraint that required target Wikipedia articles
to come from the same category. Candidate lists were augmented with link
information whenever a given anchor-text mentioned the same target
entity from at least two different Wikipedia pages.
His approach 
was 
later reimplemented~(see \S\ref{sec:comparative-results}) by \citet{Hachey2012},
who also reimplemented the earlier system of \citet{DBLP:conf/eacl/BunescuP06}.

\citet{Fader09} also generated candidates as did
\citet{DBLP:conf/emnlp/Cucerzan07}. 
They introduced prior probabilities ---
estimated from the numbers of anchors that refer to entities --- in
addition to considering the overlap between the context of a
mention and the text of the target articles.

\subsection{Wikification}
\label{sec:wikification}

Most research on wikification obviates candidate generation 
and focuses on disambiguation.
In seminal work~\citep{DBLP:conf/cikm/MihalceaC07,10.1109/MIS.2008.86}, the authors
used mentions in anchors to train a supervised~(na\"ive Bayes)
classifier.  This is the work most similar to ours. However, they did not address the
problems of building a dictionary or different methods to collect training data (cf. Section \ref{sec:variations}).
%
Wikification work continued with \citet{Milne:2008}, who combined
popularity and relatedness~(computed as the number of inlinks shared
between the context and target articles), using
several machine learning algorithms. They were the first to use the link structure of Wikipedia.
%
\citet{Kulkarni:2009:CAW:1557019.1557073} later proposed a
method that collectively wikified an entire document, by solving
a global optimization problem, using ideas from both
\citet{Milne:2008} and \citet{DBLP:conf/emnlp/Cucerzan07},
but now applied in the context of wikification. 
%

\citet{zhou-EtAl:2010:PAPERS} also built on ideas of \citet{Milne:2008} and
included, for the first time, relatedness between entities~(based on search engine log
analysis). They tested two classifiers~(binary and learning-to-rank), with
mixed results. 
\citet{DBLP:conf/acl/RatinovRDA11} utilized link structure ---
again, following \citet{Milne:2008} ---
to arrive at coherent sets of disambiguations for input
documents.  They used a ranker to select best-fitting entities.
Although anchor context was used to
compute similarity between article texts, it was not tapped for
features during classification.  Evaluation against their own
dataset 
showed that improvements over local disambiguation were small.
\citet{guo_graph-based_2011} use direct hyperlinks between the
target entity and the mentions in the context, using directly the number of
such links.

\subsection{Current NED Systems}
\label{sec:current-systems}

Given the popularity of NED, we focus this review on systems that
evaluate against TAC-KBP 2009 and 2010 data --- two of the most cited
NED datasets --- to which we will compare our own results~(see
\S\ref{sec:comparative-results}). 

In 2009, \citet{varma2009} obtained the best
results, using anchors from Wikipedia to train one
classifier per target string, combined with querying of
mention contexts against an online search engine.  
For candidate generation they used
a complex mixture of redirect links, disambiguation pages and bold text in
leading paragraphs of articles, metaphones to
capture possible spelling variations and errors, and a separate module
that tried to find acronym expansions. In addition,
they used a dynamic algorithm that returned entities matching some
(but not all) tokens of a target string. 

In 2010, \citet{citeulike:9118357} did best.
They used titles, redirects, anchors and disambiguation pages, as well
as Google searches, dynamic generation of acronym expansions, longer
mentions and inexact matching. Their system pulled in features from
both \citet{Mcnamee_2010} and \citet{Milne:2008}
to train a binary classifier.
\citet{Mcnamee_2010} had the highest-scoring submission among
systems that did not use the text in target Wikipedia articles, focusing
on the provided KB alone for building the list of candidate entities. 
A closely related system by \citet{dredze-EtAl:2010:PAPERS}
included finite-state transducers that had been trained to
recognize common abbreviations, such as ``J Miller''
for ``Jennifer Miller.'' 

\citet{zheng-EtAl:2010:NAACLHLT} 
also use a number of features that resemble those of \citet{Mcnamee_2010}.
They evaluated several learning-to-rank systems, with ListNet yielding best
results.  Unfortunately their systems were developed, trained and tested on
the same corpus, making it unclear whether their results are comparable
to those of other systems evaluated in TAC-KBP 2009.  
A similar system~\citep{zhang-EtAl:2010:PAPERS} made use of
a synthetic corpus, with unambiguous occurrences of strings
replaced by ambiguous synonyms.  Manufactured
training data was then combined with Wikipedia.
Unfortunately, since the overall system was developed and evaluated on
the same~(2009) corpus, it may have been overfitted.

\citet{han-sun:2011:ACL-HLT2011} 
proposed a generative model with three components:
popularities of entities, probabilities of strings lexicalizing an
entity, and probabilities of entire documents given a seed entity,
estimated from anchor data. They included a translation model,
learning string-entity pairs from thousands of training examples.
Although their results are highest for TAC-KBP 2009 published
to date, they may be overstated, since some parameters were tuned
using cross-validation over test data.  

Recently, \citet{Hachey2012} reimplemented three well-known NED
systems~\citep{DBLP:conf/eacl/BunescuP06,DBLP:conf/emnlp/Cucerzan07,varma2009},
combined them, and carefully analyzed the performance of candidate generation
and disambiguation components for each.
They studied contributions from a variety of available candidate sources,
including titles (of articles, redirects and disambiguation pages) and
link anchors, as well as two additional heuristics: bold in first
paragraph and hatnote templates from popular entities to corresponding
disambiguation pages.  


None of the published NED systems, except \cite{10.1109/MIS.2008.86}, uses the
context of anchors in Wikipedia to train a classifier for each
mention, as we do. In fact, among other relevant systems which have
been tested against data sets other than TAC-KBP, all use other kinds
of techniques. Given that those other systems have been evaluated on
other datasets, we will skip their description here.  As mentioned in
the introduction, most of the papers on NED describe complex systems
with many components and data sources but few ablation studies that
might help understand the contribution of each.\footnote{Again, with
  the notable exception of work by \citet{Hachey2012}, which we will
  discuss in detail~(see \S\ref{sec:comparative-results}).}  In
contrast, our system relies solely on static dictionary look-ups and
supervised classification of mentions.

\begin{figure}[t]
\includegraphics[width=\textwidth]{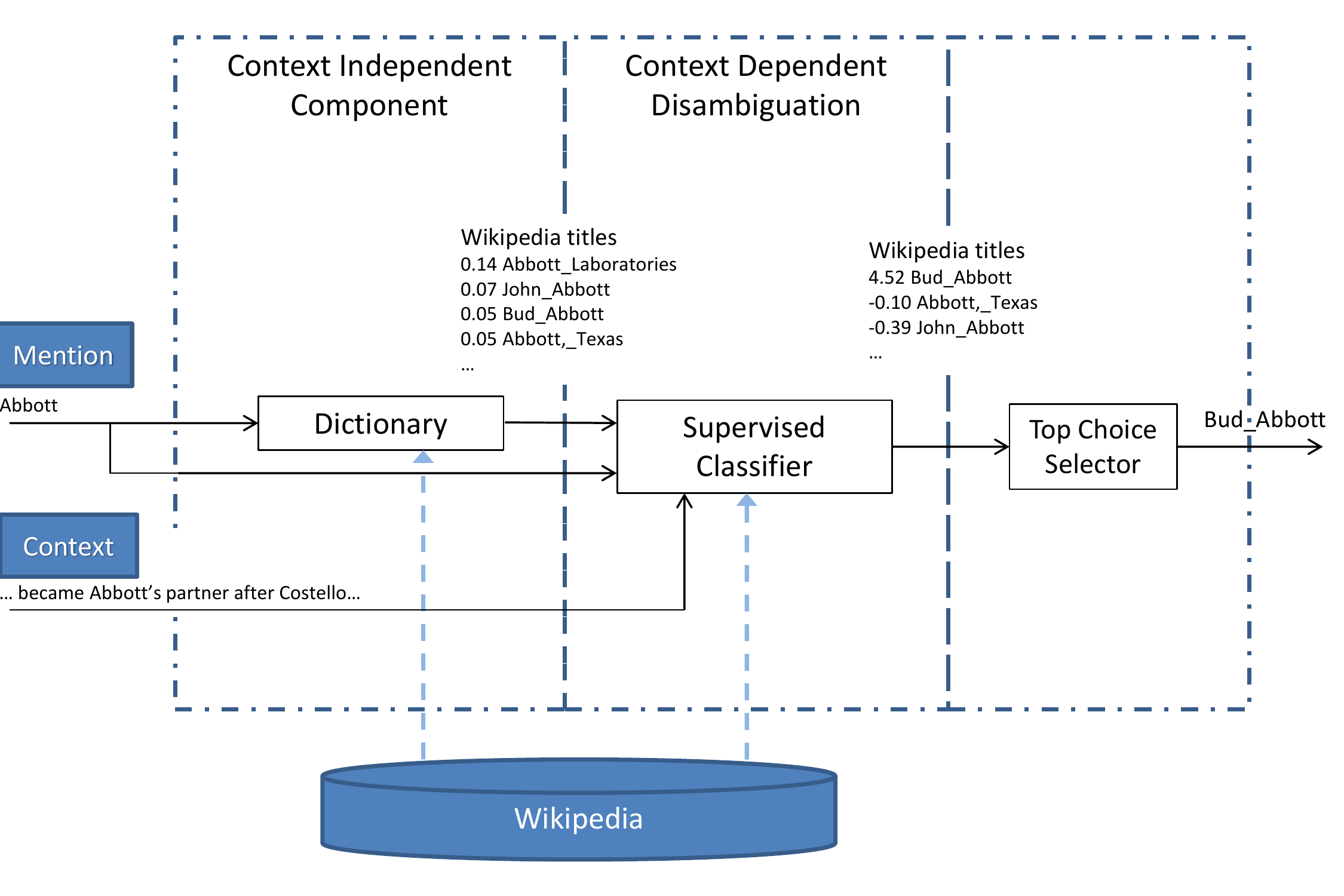}
\caption{System architecture.}\label{fig:arch}
\end{figure}

\section{System Architecture}\label{sec:architecture}

Figure~\ref{fig:arch} summarizes our approach to NED, which
is inspired by the word-expert technique, popular in mainstream WSD
systems. We take a surface text form~(mention), with the context of
its occurrence, and determine an appropriate entity~(meaning) to which
the input string may refer.  In this example, our system disambiguates
the mention ``Abbott'' in the sentence \emph{``The voice was provided by Candy
Candido, who briefly became \textbf{Abbott}'s partner after Costello had
died.''} First, a context-independent component~(the dictionary) expands
the string to a set of potentially-referent candidate entities,
ordered by popularity;
next, a context-sensitive component~(supervised classifier)
selects a candidate that seems
most appropriate for the context. 

The context-independent dictionary maps strings to ordered lists of
possible Wikipedia titles; it also provides scores ---
which are indicative of conditional
probabilities of each article given the string
--- that determine the ranking.
This (untuned) component is constructed, primarily, using
anchor-texts --- of (internal) hyper-links between English Wikipedia articles
and (external) web-links into the greater Wikipedia's pages,
covering many languages~(see
\S\ref{sec:dictionary}
for a detailed description of the dictionary and its variants).
According to our dictionary, the most probable entity for the
mention ``Abbott'' --- ignoring context --- is
\url{Abbott_Laboratories}.

From the set of candidate Wikipedia titles, a classifier
can select more appropriate entities, using not only the
mention but also surrounding text, in the
relevant document.  This (optimized) component is trained on
contexts around inter-Wikipedia links, i.e., pointers at entities
mentioned in Wikipedia articles.
We train a separate classifier for every string:
its job is to return the entity which best fits the shared mention's
context~(this and other variants of our supervised classifers
are described in \S\ref{sec:superv-disamb}).
In light of context, our system prefers the entity
\url{Bud_Abbott}.

One could view these two components as performing
(i)~candidate generation,
listing all possible entities for a mention; and
(ii)~candidate selection~\citep{Hachey2012}, although
our generation module also scores and ranks entities~(by popularity).
Following the WSD convention,
we will refer to these subsystems as (i)~the dictionary;
and (ii)~the disambiguation component.

\section{The Dictionary}\label{sec:dictionary}

The dictionary is a cornerstone component of our system, serving two
objectives.  Its primary goal is to provide a short list of candidate
referent Wikipedia articles for any string that could name an entity.
In addition, it provides a score that
quantifies the string's affinity for each candidate entity.

If the dictionary fails to recognize that a given string could refer to a
particular entity, then
our system will not be able to return that entity.  Thus, the dictionary
introduces a performance bottleneck.  We will use this fact to measure an
upper bound on our system's performance, by means of an oracle
whose choices are restricted to just the entities proposed by the dictionary.

The dictionary represents a set of weighted pairs --- an
exhaustive enumeration of all possible string-entity combinations,
with corresponding strengths of association.  We constructed this
resource from all English Wikipedia pages~(as of the March 6th, 2009 data dump)
and many references to Wikipedia from the greater web~(based on a subset of
a 2011 Google crawl).  Individual string-entity co-occurrences
were mined from several sources:

\begin{itemize}
\item Article titles that name entities, e.g., ``Robert
  Redford'' from
  \url{http://en.wikipedia.org/wiki/Robert_Redford}.\footnote{In the
    remainder of this article, we will use the following conventions:
    ``string'' for a string that can be used to name an entity (e.g.,
    ``Robert Redford'') and sufixes of English Wikipedia URLs, without the
    prefix \url{http://en.wikipedia.org/wiki/}, as short-hand for
    corresponding articles~(e.g., \url{Robert_Redford}).}
  Many title strings had to be processed, e.g., separating
  trailing parentheticals, like ``(athlete)'' in
  Figure~\ref{fig:jonathanathlete}, and underscores, from names proper.

\item Titles of pages that redirect to other Wikipedia pages, e.g.,
  ``Stanford'' for the article \url{Stanford_University}, redirected from
  the page \url{Stanford}.  

\item Titles of disambiguation pages that fan out to many similarly-named
  articles, e.g., linking
``Stanford'' to \url{Stanford_University},
  \url{Aaron_Stanford} or \url{Stanford,_Bedfordshire},
  among other possibilities, via \url{Stanford_(disambiguation)}.

\item Anchor text, e.g, we interpret the fact
  that Wikipedia editors linked the two strings
  \emph{``Charles Robert Redford''} and \emph{``Robert Redford Jr.''} to
  the article \url{Robert_Redford} as a strong indication that both
  could refer to ``Robert Redford.''  We use the number of links connecting a particular string with a specific entity as a measure of the strength of association.
\end{itemize}

Note that our dictionary spans not just named entities but also many
general topics for which there are Wikipedia articles.  Further,
it transcends Wikipedia by including anchors (i)~from the
greater web; and (ii)~to Wikipedia pages that may not (yet) exist.
For the purposes of NED,
it could make sense to discard all but the articles that correspond to
named entities.  We keep everything, however, since not all articles have
a known entity type, and because we would like to construct a resource
that is generally useful for disambiguating concepts.  Our dictionary
can disambiguate mentions directly,
simply by returning the highest-scoring entry for a given string.  Next, we describe the method used to create several
different variants of the dictionary.

\subsection{Redirects and Canonical Pages}
\label{sec:redir-canon-articl}

One of many hassles involved in building a Wikipedia-based dictionary stems
from redirects,
as it is important to separate articles that are actual
entries in Wikipedia from other place-holder pages that redirect to them.
We will use the term URL in this context to refer indiscriminately to
any article page (redirect or not), e.g., both
{\small \url{http://en.wikipedia.org/wiki/Stanford}} and
{\small \url{http://en.wikipedia.org/wiki/Stanford_University}},
where it is forwarded.

To collect all titles and URLs that refer to the same article,
we first map all such strings
to URLs using Wikipedia's canonicalization
algorithm.\footnote{http://en.wikipedia.org/wiki/Wikipedia:Canonicalization}  
We then connect any two URLs that either appear together
as an official redirection~(in a Wikipedia dump) or get
redirected at crawl-time (via HTTP status codes 3xx).
For each title and URL, we then extract a connected
component, represented by a single canonical article,
with preference given to (1)~non-redirect pages from
the Wikipedia dump; followed by (2)~official redirect pages;
and finally (3)~pages that
did not appear in any official dump.\footnote{This process can be
particularly complicated when reconciling snapshots of Wikipedia
taken at different times~(i.e., an official dump and a web crawl),
since sources and targets of redirects can switch roles over time,
yielding cycles.}
Within each preference category,
we resolved ties lexicographically.  Figure~\ref{fig:remapping}
shows a simple example clustering of URLs, with a canonical article.

\begin{figure}
{\footnotesize\begin{center}\begin{tabular}{ll}
\url{Route_102_(Virginia_pre-1933)}&
\url{State_Route_102_(Virginia_1928)} \\
\url{State_Route_102_(Virginia_1928-1933)} &
\url{State_Route_102_(Virginia_pre-1933)} \\
\url{State_Route_63_(Virginia_1933)} &
\url{State_Route_63_(Virginia_1933-1946)} \\
\url{State_Route_63_(Virginia_1940)} &
\url{State_Route_63_(Virginia_pre-1946)} \\
\url{State_Route_758_(Lee_County,_Virginia)} &
\textbf{\url{Virginia_State_Route_758_(Lee_County)}}
\end{tabular}\end{center}}
\caption{A connected component of URLs that redirect to \textbf{\protect\url{Virginia_State_Route_758_(Lee_County)}}, which itself does not redirect further and is therefore selected as the canonical article for the entire cluster.}
\label{fig:remapping}
\end{figure}

\subsection{The Core and Other Dictionaries}

We created a core dictionary by extracting strings from titles
of canonical articles, redirects and disambiguation pages, as
well as the referring anchor text encountered in both Wikipedia
and the Google crawl. In all
cases, we paired a string with the canonical article of the
respective cluster.

Our core dictionary maps strings to sorted lists of 
Wikipedia articles, with associated scores. These scores are
computed from the occurrence frequencies of the anchor texts.
For a given string-article pair, where the string has been
observed as the anchor-text of a total of $y$ inter-Wikipedia
--- and $v$ external --- links, of which $x$ (and,
respectively, $u$) pointed to a page that is represented by the
canonical article in the pair, we set the pair's score to
be $(x+u)/(y+v)$.


We call this dictionary exact (\textbf{EXCT}), as it matches precisely
the raw strings found using the methods outlined above.  For example,
Figure \ref{fig:exct} shows all eight articles that have been referred
to by the string \emph{``Hank Williams.''}
Note that this dictionary does no filtering:
removal of undesired target Wikipedia pages
(such as disambiguations) will be done at a later
stage~(see Section \ref{sec:using-dict-disamb}). 

\begin{figure}
{\footnotesize\begin{center}\begin{tabular}{lll}
0.9976 & \url{Hank_Williams}
\phantom{HACKHACKHACKHACK}                          & {\tt{w:756/758 }}\phantom{ }{\tt{ W:936/938}} \\ 
0.0012 & \url{Your_Cheatin'_Heart}                  & \phantom{\tt{w:767/758 }}\phantom{ }{\tt{ W:\phantom{93}2/938}} \\                    
0.0006 & \url{Hank_Williams_(Clickradio_CEO)}       & {\tt{w:\phantom{75}1/758}} \\  
0.0006 & \url{Hank_Williams_(basketball)}           & {\tt{w:\phantom{75}1/758}} \\  
0      & \url{Hank_Williams,_Jr.}               \\
0      & \url{Hank_Williams_(disambiguation)}   \\
0      & \url{Hank_Williams_First_Nation}       \\
0      & \url{Hank_Williams_III}                \\
\end{tabular}\end{center}}
\caption{Sample from the \textbf{EXCT} dictionary,
listing all articles and scores for the string
\emph{``Hank Williams.''}
\protect\linebreak
Final column(s) report counts from Wikipedia
({\tt{w}}:$x/y$) and the web crawl ({\tt{W}}:$u/v$), where available.}
\label{fig:exct}
\end{figure}

\subsubsection{Aggregating String Variants}

In addition to exact string lookups, we also adopt two
less strict views of the core dictionary.  They capture
string variants whose lower-cased normalized
forms are either the same~(the \textbf{LNRM} dictionary) or 
close~(the \textbf{FUZZ} dictionary) by Levenshtein edit-distance to that of the string
being queried.  In both cases, an
incoming string now matches a \emph{set} of keys (strings) in the
dictionary, whose lists of scored articles are then merged,
as follows: given $n$ articles, with scores $a_i/b_i$,
their aggregate score is also a ratio, $\left(\sum_{i=1}^n a_i\right) /
\left(\sum_{i=1}^n b_i\right)$.


We form the lower-cased normalized variant $l(s)$ of a string $s$ by
canonicalizing Unicode characters, eliminating diacritics,
lower-casing and discarding any resulting ASCII-range characters
that are not alpha-numeric.  If what remains is the empty string,
then $s$ maps to no keys; otherwise, $s$ matches all keys $k$ such that
$l(s)=l(k)$, with the exception of $k=s$, to exclude the original key~(which
is already covered by \textbf{EXCT}).
Figure \ref{fig:lnrm} shows a subset of the
\textbf{LNRM} dictionary, with combined contributions
from strings $k$ that are similar to (but different from) $s$.

\begin{figure}
{\footnotesize\begin{center}\begin{tabular}{lll}
0.9524  & \url{Hank_Williams}
\phantom{HACKHACKHACKHACK}                           & \phantom{\tt{w:756/758 }}\phantom{ }{\tt{ W:20/21 }} \\ 
0.0476  & \url{I'm_So_Lonesome_I_Could_Cry}          & \phantom{\tt{w:756/758 }}\phantom{ }{\tt{ W:\phantom{2}1/21}} \\       
0       & \url{Hank_Williams_(Clickradio_CEO)}  \\
0       & \url{Hank_Williams_(basketball)}      \\
0       & \url{Hank_Williams_(disambiguation)}    
&\phantom{\tt{w:756/758 }}\phantom{ }\phantom{\tt{ W:936/938}}
\end{tabular}\end{center}}
\caption{Sample from the \textbf{LNRM} dictionary (note
that contributions already in \textbf{EXCT} are excluded),
for strings with forms like
  $l(\mbox{\emph{Hank}}\mbox{
  }\mbox{\emph{Williams}})=\mbox{\emph{hankwilliams}}
  =l(\mbox{\emph{HANK}}\mbox{
  }\mbox{\emph{WILLIAMS}})=l(\mbox{\emph{hank}}\mbox{
  }\mbox{\emph{williams}})$, etc.}
\label{fig:lnrm}
\end{figure}

We define the \textbf{FUZZ} dictionary via a metric,
$d(s,s')$: the byte-level UTF-8 Levenshtein edit-distance between
strings $s$ and $s'$.  If $l(s)$ is empty, then $s$ maps to no keys
once again; otherwise, $s$ matches all keys $k$, whose $l(k)$ is also
not empty, that minimize $d(l(s),l(k))>0$.
This approach excludes not only $k=s$ but also $l(k)=l(s)$,
already covered by \textbf{LNRM}. For example,
for the string
\emph{Hank Williams}, there exist keys
whose signature is exactly one byte away, including
\emph{Tank Williams}, \emph{Hanks Williams},
\emph{hankwilliam}, and so forth.  Figure \ref{fig:fuzz} shows the three
articles ---
two of them already
discovered by both \textbf{EXCT} and \textbf{LNRM} dictionaries
--- found via this fuzzy match.

\begin{figure}
{\footnotesize\begin{center}\begin{tabular}{lll}
0.6316 & \url{Tank_Williams}        & {\tt{w:12/12}} \\ 
0.3158 & \url{Hank_Williams}
\phantom{HACKHACKHACKHACK}          & \phantom{\tt{w:756/758 }}\phantom{ }{\tt{ W:6/7}}\phantom{\tt{9338}} \\      
0.0526 & \url{Your_Cheatin'_Heart}  & \phantom{\tt{w:756/758 }}\phantom{ }{\tt{ W:1/7}} \\
\end{tabular}\end{center}}
\caption{Sample from the \textbf{FUZZ} dictionary
 (note that contributions already in \textbf{EXCT}
  or \textbf{LNRM} are excluded).}
\label{fig:fuzz}
\end{figure}

\subsubsection{Querying Search Engines}

In addition to constructing our own custom dictionaries,
we also experimented with
emulating a lookup by (twice) querying the Google search
engine,\footnote{\url{http://www.google.com/}}
which we call the \textbf{GOOG} dictionary, on the fly.
Our query pairs consisted of the raw target string, \url{<str>}
--- and also a separately-issued phrase-query, ``\url{<str>}'' ---
combined with a restriction, \url{site:en.wikipedia.org}.
We kept only the returned URLs that begin
with \url{http://en.wikipedia.org/wiki/},
using their (sums of) inverse ranks as scores.
Although this method cannot be used effectively to find all strings that may
name an entity, given any string, the dictionary portion for
that specific string can be filled in dynamically, on demand. 



\subsection{Disambiguating with Dictionaries}
\label{sec:using-dict-disamb}

For any input string, one could simply use a dictionary to
select a highest-scoring entity.  Our core dictionary's scores
already capture the popularity of each entity given a string,
approximated by the frequency of their pairing in anchor texts.
The Google dictionary may, in addition, reflect
relatedness to the text of the corresponding Wikipedia article,
according to more sophisticated ranking algorithms.
Such approaches would be closely related to the MFS heuristic in WSD
and could also be viewed as context-free disambiguation, since they do
not take into account the context of the mention, always returning
the same article, regardless of the surrounding text.

A set of dictionaries, like ours, can be abstracted away behind a single
dictionary interface, to be used by downstream components.  Internally,
dictionaries could be combined using a variety of strategies.  We
suggest a backoff approach that checks the dictionaries
in order of precision, and also a heuristic that discards
very unlikely candidates.  Other strategies, such as using 
decision trees or a weighted combination of the dictionaries, are also possible.
But many of these strategies would involve parameter tuning (e.g.,
learning the weights). A thorough exploration of more optimal techniques for
combining dictionaries lies outside of the present paper's scope.

Our first strategy is a \textbf{cascade of dictionaries}:
for a given string, it provides a ranked list of entities
by consulting the dictionaries
in order of their precision.  It starts from \textbf{EXCT},
which is most precise, backing off to the rest.
We consider two specific cascades (in remaining sections,
all references to the \textbf{LNRM} dictionary
correspond to the \textbf{LNRM} \emph{cascade},
and similarly for \textbf{FUZZ}):
\begin{compactitem}
\item the \textbf{LNRM} cascade first checks the \textbf{EXCT}
  dictionary, returning its associated entities and scores if the
  string has an entry there, and defaulting to \textbf{LNRM} dictionary's
  results if not;
\item the \textbf{FUZZ} cascade also first checks the \textbf{EXCT}
  dictionary, backs off to the \textbf{LNRM} dictionary in case of
  a miss, but then finally defaults to the contents from
  the \textbf{FUZZ} dictionary.
\end{compactitem}

Our second strategy is a \textbf{heuristic combination} (\textbf{HEUR}):
it combines suggestions from the original (\textbf{EXCT}, \textbf{LNRM},
and \textbf{FUZZ}) dictionaries while also filtering out some of the noise.
Its goal is to retain good suggestions without drowning in
obviously bad choices: since many titles suggested by
the \textbf{FUZZ} dictionary are noisy, we only include
those for which there is additional evidence of
relevance (for instance, if the suggestion is an acronym for the string).  
Similarly, if a string has been used to link to an article only a few times,
the connection between them may not be reliable, calling for more evidence
before the entity could be accepted as a possible referent for the string.
Naturally, we also discard articles that are clearly not real entities ---
such as disambiguation pages, ``list of'' pages, and pages of dates ---
using additional features collected with the core dictionary.
Table \ref{table:heurdict} summarizes our complete list of heuristics,
which was finalized by consulting the {\em{news}} subset of the 2010
TAC-KBP development dataset~(see Section \ref{sec:tac-kbp-dataset}). 

Heuristic combinations can yield dictionaries that are
considerably smaller than full cascades.
For example, for the string \emph{ABC}, the \textbf{EXCT} dictionary offers 191
options, the \textbf{LNRM} cascade 253, and the \textbf{FUZZ} cascade 3,527.
But with the above-mentioned filters, the number of candidate Wikipedia
titles can be reduced to just 110.  Although our
heuristics could be applied to any dictionary or cascade, in the remainder
of the paper we will use the short-hand \textbf{HEUR} to refer to
the larger \textbf{FUZZ} cascade, followed by the application of these
heuristic filtering rules.

Inevitably, \textbf{HEUR} will sometimes discard a correct mapping
in favor of inferior choices.  For instance, the feasible suggestion
\emph{Angela Dorothea Kasner} $\to$ \url{Angela_Merkel} gets dropped,
but the undesirable association \emph{Angela Merkel}
$\to$ \url{German_federal_election,_2005} is kept.
Nevertheless, with a reduced set of candidates,
it will be easier for supervised classifiers to narrow down remaining options
based on a document's context, down the road~(see \S\ref{sec:superv-disamb}).

\begin{savenotes}
\begin{table}
\begin{center}
\begin{tabular}{r|p{5.5cm}|l}
& \em{Rule} & \em{Example} \\
\hline
\scriptsize{1} & \footnotesize{Discard disambiguation pages.} 
&
\footnotesize{Discard:} \hfill \scriptsize{\tt{*} $\to$ \tt{Hank\_Williams\_(disambiguation)}}
\\
\scriptsize{2} & \footnotesize{Discard date pages.} 
&
\footnotesize{Discard:} \hfill \scriptsize{\tt{*} $\to$ \tt{2000}}
\\
\scriptsize{3} & \footnotesize{Discard list-of pages.}
& \footnotesize{Discard:} \hfill \scriptsize{\tt{*} $\to$ \tt{List\_of\_cheeses}}
\\
\scriptsize{4} & \footnotesize{Discard pages only suggested by \textbf{FUZZ}, unless:} 
& \footnotesize{Discard:} \hfill \scriptsize{\emph{MND} $\to$ \tt{MNW}} 
\\
&
$\cdot$ \footnotesize{string and title could be an acronym pair;}\footnotemark
& \footnotesize{$\;\;\;\;$Keep:} \hfill \scriptsize{\emph{NDMC} $\to$ \tt{National\_Defense\_Medical\_Center}}
\\
& 
$\cdot$ \footnotesize{string is a substring of the title;}
& \footnotesize{$\;\;\;\;$Keep:} \hfill \scriptsize{\emph{DeLorean Motor} $\to$ \tt{DeLorean\_Motor\_Company}}
\\
&
$\cdot$ \footnotesize{string is very similar to the title.}\footnotemark
& \footnotesize{$\;\;\;\;$Keep:} \hfill \scriptsize{\emph{Chunghua Telecom} $\to$ \tt{Chunghwa\_Telecom}}
\\
\scriptsize{5} & \footnotesize{Discard articles supported by few links,\footnotemark unless:}
& \footnotesize{Discard:} \hfill \scriptsize{\emph{Washington} $\to$ \tt{Tacoma,\_Washington}}
\\
&
$\cdot$ \footnotesize{article may disambiguate the
string;} 
& \footnotesize{$\;\;\;\;$Keep:} \hfill \scriptsize{\emph{CNS} $\to$ \tt{Szekler\_National\_Council}}
\\ 
&
$\cdot$ \footnotesize{string is the title of the page.} 
& \footnotesize{$\;\;\;\;$Keep:} \hfill \scriptsize{\emph{Chunghwa Telecom} $\to$ \tt{Chunghwa\_Telecom}}
\\
\end{tabular}
\end{center}
\caption{Rules of the heuristic combination dictionary~(\textbf{HEUR}).}
\label{table:heurdict}
\end{table}
\end{savenotes}
\addtocounter{footnote}{-3}
\addtocounter{footnote}{1}\footnotetext{I.e., either (a)~the string is an acronym for the title;
or (b)~the title is an acronym for the string.}
\addtocounter{footnote}{1}\footnotetext{I.e., either (a)~the strings are the same;
or (b)~both strings have length less than or equal to six, with an edit distance exactly equal to one;
or (c)~the ratio between edit distance and string length is less than or equal to 0.1.}
\addtocounter{footnote}{1}\footnotetext{I.e., if (a) the number of total links to the page (both inside Wikipedia
and from the external web) is no more than ten; or (b)~the number
of times the string links to the page is no more than one; or
(c)~the score is no more than 0.001.} 

\section{Supervised Disambiguation}
\label{sec:superv-disamb}

The large number of naturally-occurring links pointing at entities in Wikipedia
makes it possible to gather richly-annotated data automatically, to
augment context-free disambiguation provided by the
raw frequency information the dictionaries.
In this respect, NED differs from many other applications in natural
language processing and information retrieval, where most available
input data is not already pre-labeled,
and substantial resources are
devoted to manual annotation efforts. 

\begin{figure}

\begin{center}\footnotesize\tt\begin{tabular}{l}
On  February 27,  2004,  SuperFerry 14  was  bombed  by  the \underline{Abu Sa}yy\underline{af} terrorists \\
killing 116 people . It was considered as the worst terrorist attack ...
\end{tabular}\end{center}

\begin{center}\scriptsize\begin{tabular}{rlc}
  \multicolumn{1}{r}{\it{anchor text}} &
    & {\tt{Abu\_Sayyaf}} \\
\\
  \multicolumn{1}{r}{\it{lemmas in the span}} &
    & {\tt{ terrorist}} \\
    && {\tt{ kill}} \\
    && {\tt{...}} \\
\\
  \multicolumn{1}{r}{\it{lemma for N/V/A}}  &
    & {\tt{ be}} \\
    {\it{in a 4 token window}} && {\tt{ bomb}} \\
    {\it{around the anchor text}} && {\tt{ kill}} \\
    && {\tt{ people}} \\
    && {\tt{ terrorist}} \\
\\
  {\it{lemma and word for N/V/A}}
    & noun (lemma) & {\tt{ SuperFerry}} \\
  {\it{before the anchor text}}  & noun (word) & {\tt{ SuperFerry}} \\
    & verb (lemma) & {\tt{ bomb}} \\
    & verb (word)  & {\tt{  bombed}} \\
\\
  {\it{lemma and word for N/V/A}}
    & adjective (lemma) & {\tt{ bad}}  \\
  {\it{after the anchor text}}  & adjective (word) & {\tt{ worst}} \\
    & noun (lemma) & {\tt{ terrorist}} \\
    & noun (word) & {\tt{ terrorists}} \\
    & verb (lemma) & {\tt{ kill}} \\
    & verb (word) & {\tt{ killing}} \\
\\
   {\it{bigrams around anchor text}}
    & lemma before & {\tt{ the Abu\_Sayyaf}} \\
    & lemma after & {\tt{ Abu\_Sayyaf terrorist}} \\
    & POS before & {\tt{ DT J}} \\
    & POS after & {\tt{ J N2}} \\
    & word before & {\tt{ the Abu\_Sayyaf}} \\
    & word after & {\tt{ Abu\_Sayyaf terrorist}} \\
\\
   {\it{trigrams around anchor text}}
    & lemma before & {\tt{ by the Abu\_Sayyaf}} \\
    & lemma around  & {\tt{ the Abu\_Sayyaf terrorist}} \\
    & lemma after & {\tt{ Abu\_Sayyaf terrorist kill}} \\
    & POS before & {\tt{ P-ACP DT J}} \\
    & POS around & {\tt{ J N2 VVG}} \\
    & POS after & {\tt{ DT J N2}} \\
    & word before & {\tt{ by the Abu\_Sayyaf}} \\
    & word around & {\tt{ the Abu\_Sayyaf terrorists}} \\
    & word after & {\tt{ Abu\_Sayyaf terrorists killing}} \\
\end{tabular}\end{center}

\caption{Example training context and features extracted from Wikipedia's article for {\small\protect\url{SuperFerry}}.}
\label{fig:features}
\end{figure}

\subsection{Core Method}
\label{sec:core-method}

We followed a mainstream WSD approach~(word-expert), training supervised classifiers for all target strings, as follows:
For every string in the dictionary, we first identify the entities to which it may refer. We then gather all example spans from Wikipedia articles that contain links to any of these entities.  To ensure that our training data is natural language (and not, e.g., lists or tables),
we only include text marked as paragraphs (i.e., enclosed between HTML tags \texttt{<P>} and \texttt{</P>}).  The relevant training subset for a target string then consists of example contexts with anchor texts containing the string.\footnote{The target string
is a substring of the anchor text after case normalization.} We take spans of up to 100 tokens to the left --- and another 100 to
the right --- of a link to be the contexts.  Figure \ref{fig:features} shows one such sample
training instance in detail.

Since the \textbf{EXCT} dictionary often provides too few examples, we used the \textbf{LNRM} cascade
as our source dictionary~(with remapped canonical articles --- see \S\ref{sec:redir-canon-articl}).
Given this training data, we applied standard machine learning
techniques to perform supervised disambiguation of entities.  
We trained a multi-class classifier for
each target string.  Then, given a mention of the target string in the
test data, we applied its classifier to the context of the mention,
and returned the corresponding article. We did not construct
classifiers for strings whose training data maps to a unique entity.
Instead, in those cases, a default classifier falls back to \textbf{LNRM} cascade's output.

From each context, we extracted features~(see Figure \ref{fig:features})
commonly used for supervised classification in
the WSD setting~\citep{agirre-lopezdelacalle:2007:SemEval-2007,zhong-ng:2010:Demos}:
\begin{compactitem}
\item the anchor text;
\item the unordered set of lemmas in the span;
\item lemma for noun/verb/adjective in a four-token window around the
  anchor text;
\item lemma/word for noun/verb/adjective before and after the anchor
  text;
\item word/lemma/part-of-speech bigram and trigrams including the
  anchor text.
\end{compactitem}

\subsection{Variations}
\label{sec:variations}

Over the course of developing our system,
we tested several variations of the core algorithm:\linebreak
\textbf{Classifier}: We tried maximum entropy models~(MAXENT) and support vector machines~(SVM).\linebreak
\textbf{Dictionary}: A dictionary influences supervised classification in two places.
First, when building the training data, to filter example spans selected for training.
And second, as a backup ranker, for cases when a classifier is not trained, due to a lack of examples.
In both the filtering stage and the back-off stage, we compared using the \textbf{HEUR} dictionary
in place of the \textbf{LNRM} cascade.\\
\textbf{Span}: In addition to training with contexts of
(up to) 100 tokens to the left and right of a string, we
also tried single-sentence and full-paragraph
spans~(the \textbf{100}, \textbf{SENT} and \textbf{PARA} variants).\linebreak
\textbf{Match}: When gathering examples for a
target string, we made sure that the anchor text contains this
string~(the \textbf{LEX} default). Alternatively, we could allow
additional examples, ignoring anchor text mismatch~(the \textbf{SENSE} variant):
given the entities that a dictionary lists for the target string, we
include as training examples all contexts that apply to these entities,
regardless of their anchor text. 
In this variant, the target string is simply treated as another feature
by the classifier.  If a test example's string does not match any of the anchor text seen in training, then features that include the target string (i.e., its unigram, bigram, and trigram features) will not fire.  Classification will then depend on features describing the rest of the context: a classifier could still give a high score, but only if surrounding words of a span carry a strong enough signal for an entity.
This approach may allow us to classify aliases for which there isn't exact training data,  
provided that our filtering dictionary yields a precise list of potential entities corresponding to a target string.

\section{Datasets for NED}
\label{sec:dataset}

Evaluation of NED systems requires manually annotated data.
Although many corpora have been introduced in various papers,
we decided to focus on the earlier datasets developped
 for the entity linking task of the
knowledge-base population~(KBP) track at
Text Analysis Conferences~(TAC),\footnote{\url{http://www.nist.gov/tac/}}
which have been running annually each year since 2009~\citep{McNameeAndDang09,JiEtAl10,ji-grishman:2011:ACL-HLT2011}.

The TAC-KBP evaluation
focuses on three main types of named entities: persons~(PER), organizations~(ORG) and
geo-political entities~(GPE).  Given a set of hand-selected mentions of
entities --- and documents containing these strings --- the task is to determine
which knowledge-base instance, if any, corresponds to each named entity string.
The knowledge-base~(KB) is derived from a subset of Wikipedia.  Mentions
are chosen among occurrences in a collection of 1,286,609 newswire documents and
490,596 web-pages.  The tasks' organizers have released substantial amounts of
development and test data, as well as standardized evaluation
software.

There are several reasons why we chose to evaluate against TAC-KBP data:
(i)~it consists of named entity mentions from two genres (news articles and pages crawled off the web); (ii)~it focuses on
several taget entity mentions, making it well suited to our word-expert approach; and
(iii)~its high number of participating systems --- and subsequent
publications --- provide an informative setting for comparing
state-of-the-art techniques. Among TAC-KBP datasets, those from  2009 and 2010
attract the largest number of papers~\citep{varma2009,Mcnamee_2010,citeulike:9118357,zheng-EtAl:2010:NAACLHLT,dredze-EtAl:2010:PAPERS,zhang-EtAl:2010:PAPERS,han-sun:2011:ACL-HLT2011,ploch:2011:SS,chen-ji:2011:EMNLP,gottipati-jiang:2011:EMNLP,Hachey2012,han-sun:2012:EMNLP-CoNLL}.
In the future we would like to extend our work to other
datasets which include all mentions in full documents~\citep{hoffart-EtAl:2011:EMNLP}.


\subsection{The TAC-KBP Dataset}
\label{sec:tac-kbp-dataset}

\begin{figure}[t]{\scriptsize        
\begin{lstlisting}[language=xml,basicstyle=\ttfamily,frame=single]
<entity wiki_title="Mike_Quigley_(footballer)" type="PER" 
    id="E0000001" name="Mike Quigley (footballer)">
<facts class="Infobox Football biography">
<fact name="playername">Mike Quigley</fact>
<fact name="fullname">Michael Anthony Joseph Quigley</fact>
<fact name="dateofbirth">October 2, 1970 (1970-10-02) (age_38)</fact>
<fact name="cityofbirth"><link entity_id="E0467057">Manchester</link></fact>
<fact name="countryofbirth"><link entity_id="E0145816">England</link></fact>
<fact name="position"><link>Midfielder</link></fact>
</facts>
<wiki_text><![CDATA[Mike Quigley (footballer)
Mike Quigley (born 2 October 1970) is an English football midfielder.
]]></wiki_text>
</entity>
\end{lstlisting}}
\caption{Example of a KB person entity from the TAC-KBP dataset. }
\label{fig:kbexample}
\end{figure}

\begin{figure}[t]{\scriptsize
\begin{lstlisting}[language=xml,basicstyle=\ttfamily,frame=single]
<query id="EL55">
<name>ABC</name>
<docid>AFP_ENG_20070104.0533.LDC2009T13</docid>
</query>
\end{lstlisting}}
\caption{Example of a query with document ID and entity name from the TAC-KBP dataset. }
\label{fig:queryexample}
\end{figure}

The TAC-KBP exercise provides an inventory of target named entities, based
on a subset of Wikipedia articles that had info-boxes in October of 2008.
This KB contains 818,741
entities~(a.k.a. KB instances), each marked with (i)~its name (a string);
(ii)~the assigned entity type~(one of PER, ORG or GPE);
(iii)~a KB instance ID (a unique identifier, such as {\tt{E001}});
(iv)~the set of info-box slot names and values from the corresponding
Wikipedia page; and
(v)~the text of that Wikipedia page.  Figure \ref{fig:kbexample}
shows a sample KB entry for a person, whose entity derives from
the Wikipedia article identified by the URL \url{Mike_Quigley_(footballer)}.

Given a query that consists of a string and a document ID~(see Figure \ref{fig:queryexample}), the task is to
determine the knowledge-base entity to which that document's string refers
(or to establish that the entity is not present in the
reference KB).  The document provides context which may be useful
in disambiguating the string.  A referent entity will generally
occur in multiple TAC-KBP queries, under different surface name
variants and in different documents.  Because of possible
auto-correlations, official rules stipulate that queries must be processed
independently of one another.  As expected, some entities share
confusable names (e.g., the string \emph{Stanford} refers to a
university, its town and founder, among many other possibilities).
For each query, a system must return a KB ID (or NIL when there
is no corresponding KB-instance). All of the queries had been tagged by a team 
of annotators. Inter-annotator agreement is high for
organizations~(ORG: 92.98\%) and people~(PER: 91.53\%),
and somewhat lower for remaining entities~(GPE: 87.5\%).

TAC-KBP has been running and releasing development and test
data each year since 2009. It attracted 13 teams in the first year and 16 teams in
2010.  We use the same data split as in the 2010 task.  For the dev-set, all \emph{news} examples come
from the 2009 test-set, and \emph{web} examples are exclusively from
the 2010 training set; the 2010 test-set contains both types. Table \ref{tab:dataset-types} gives a
break-down of our development and evaluation sets by the three entity
types (PER, ORG, GPE).  Note
that \emph{news} samples in the dev-set are especially skewed toward
ORG, with approximately five times as many examples as either PER or
GPE; the rest of the data is perfectly balanced.


\begin{table}[t]
  \begin{center}
    {\small\begin{tabular}{r|r|r|r|r|l}
      \cline{3-5}
      \multicolumn{1}{c}{\it{}}
      & \multicolumn{1}{r|}{\it{Total}}
      & \multicolumn{1}{r|}{PER}
      & \multicolumn{1}{r|}{ORG}
      & \multicolumn{1}{r|}{GPE} \\
      \cline{2-5}
      & 1,500 &   500 &    500 &  500 &{\em{web \hfill (2010 train)}} \\
      \cdashline{2-2}\cdashline{3-3}\cdashline{4-4}\cdashline{5-5}\cdashline{6-6}
      & 3,904 &   627 &  2,710 &  567 &{\em{news \hfill (2009 test)}} \\
      \cline{2-5}
      {\it{Development}}  & 5,404 & 1,127 & 3,210 & 1,067 \\
      \cline{1-5}
      {\it{Evaluation}}  & 2,250 & 750 & 750 & 750 \\
      \cline{2-5}
      &  1,500 & 500 &  500 &  500 &{\em{news \hfill (2010 test)}} \\
      \cdashline{2-2}\cdashline{3-3}\cdashline{4-4}\cdashline{5-5}\cdashline{6-6}
      &    750 & 250 &  250 &  250 &{\em{web \hfill (2010 test)}} \\
      \cline{2-5}
    \end{tabular}}
  \end{center}
  \caption{Number of examples in the development and evaluation datasets, broken down by genre and entity type. We explicitly mention the relation to the 2009 and 2010 TAC KBP datasets. }
  \label{tab:dataset-types}
\end{table}




Since the inventory of entities used in TAC-KBP is incomplete, it is
possible for queries to refer to unlisted entities.  In such cases,
human annotators tagged the mentions as NILs. Note that dthe
development and test datasets a large number of NIL mentions, 49\% and
55\%, respectively. We evaluate on non-NIL mentions.


\begin{table}[t]
  \begin{center}
    {\small\begin{tabular}{r|r|r|r|r|r|l}
      \multicolumn{1}{c}{\it{}}
      & \multicolumn{1}{c|}{\it{}}
      & \multicolumn{1}{r|}{\it{No Entities}}
      & \multicolumn{1}{r|}{\it{Single Entity}}
      & \multicolumn{2}{c}{\it{Multiple Entities}} \\
      \cline{5-6}
      \multicolumn{1}{c}{\it{}}
      & \multicolumn{1}{r|}{\it{Unique Strings}}
      & \multicolumn{1}{r|}{{\it{(}}NIL{\it{s)}}}
      & \multicolumn{1}{r|}{\it{(Monosemous)}}
      & \multicolumn{1}{r|}{\it{(Polysemous)}}
      & \multicolumn{1}{r|}{{\it{Ambiguity}}} \\
      \cline{2-6}
      {\it{Development}} & 1,162 & 462 & 488 & 112 & 2.34 \\
      \cline{1-6}
      {\it{Evaluation}}  &   752 & 366 & 317 &  69 & 2.17\\
      \cline{2-6}
    \end{tabular}}
  \end{center}
  \caption{Ambiguity of target strings in the development and evaluation datasets, according to the gold standard.}
  \label{tab:ambiguity}
\end{table}

\begin{table}[t]
  \begin{center}
    \begin{tabular}{r|r|rr|}
     \multicolumn{1}{c}{}
   & \multicolumn{1}{c|}{\emph{Words}} & \multicolumn{2}{c}{\emph{Ambiguity}} \\ \cline{3-4}
     \multicolumn{1}{c}{}
   & \multicolumn{1}{c|}{\emph{(Polysemous)}} & \emph{GS}   & \emph{Dictionary} \\ \cline{2-4}
\emph{Nouns}      & 20    & 5.05 & 5.80 \\      
\emph{Verbs}      & 32    & 4.56 & 6.31 \\      
\emph{Adjectives} & 5     & 6.20 & 10.20 \\      \cline{1-4}
\emph{Total}      & 57    & 4.88 & 6.47 \\      \cline{2-4}
\end{tabular}
\end{center}
\caption{Ambiguity for Senseval-3 \emph{lexical sample}~(a popular WSD
  dataset), including all senses covered in the gold standard (GS), as
  well as any senses attested by the dictionary; note that all of the
  words are polysemous.}
\label{fig:senseval3}
\end{table}

\subsection{Ambiguity}
\label{sec:ambiguity}

Some simple measurements can give an idea as to the difficulty of a problem or dataset.
In our case, one such quantity is a mention's \emph{ambiguity}: a lower bound on
the number of different KB-instances to which its string can refer.
In WSD, an equivalent notion would be the concept of \emph{polysemy}.
In our dictionary, for example, the string \emph{ICNC} covers four distinct entities in Wikipedia.

We estimated ambiguity as follows:
For each target string in a dataset, we counted the number
of different KB-instances associated to it by human annotators.
Table~\ref{tab:ambiguity} reports ambiguities in both
development and evaluation sets.  The section of the dev-set,
for instance, comprises 1,162 unique target strings (types).  Of those,
462 were tagged with NIL, as no referent entity was deemed
appropriate in the KB; 488 strings had all their mentions tagged with the
same entity; and only 112 had been tagged with multiple KB-instances
--- an average of 2.34 entities per ambiguous string.
Of course, this calculation grossly underestimates actual polysemy, as the
number of potential articles to which a string could refer may far exceed those
found in our collection by annotators.  As mentioned in the previous section,
populating a complete list of entities that could have been invoked is part of the task,
and as such, still an open question.  However, many target strings tend
to be highly skewed, in the sense that they usually refer to one
specific entity.  This fact is reflected by the large number of
strings which refer to a single entity in the gold standard sample.

We will show that actual polysemy is much higher~(e.g.,
according to our dictionaries --- see \S\ref{sec:ambig-synonymy-dict}),
since many of the possible entities do not appear in standard datasets.
This fact further differentiates NED from typical WSD settings, where most senses
can be found in a gold standard.  Table~\ref{fig:senseval3} shows polysemy 
as average number of senses for a WSD set-up~(Senseval-3),
with respect to both gold-standard data and a dictionary, which happen to be
fairly similar.  Yet another point of contrast is that many mentions in NED refer to only
a single entity, whereas in WSD all 57 target words are polysemous
(with multiple senses attested by the gold standard).

\subsection{Synonymy}
\label{sec:synonymy}

Another quantity that sheds light on the complexity of a disambiguation task,
in a manner that is complementary to ambiguity, is the number of different
strings that can be used to name a particular entity: its \emph{synonymy},
in WSD terms.  As before, tallying all unique strings that may refer to a
given entity is an open problem.  Table~\ref{tab:synonymy}
tabulates the gold standard's synonymy~(i.e.,
the number of strings found in the document collection by the annotators)
for each entity in the KB.  Most entities are associated with only a single
string, especially in the evaluation sets.  This could be an artifact of how
the organizers constructed the test data, since their
procedure was to first choose ambiguous strings and then find documents
containing different meanings (entities) for these mentions --- as
opposed to first choosing entities and then querying string
variants of names. The annotators thus did not search for
alternative lexicalizations (synonyms) to be used as query
strings. As was the case with ambiguity, actual synonymy is much higher~(see
\S\ref{sec:ambig-synonymy-dict}, in which we compute similar figures to
analyze some of the dictionaries used by our systems).

\begin{table}[t]
  \begin{center}
    {\small\begin{tabular}{r|r|rr|rr|r|l}
      \multicolumn{1}{c}{\it{}}
      & \multicolumn{1}{r|}{\it{Unique Entities}}
      & \multicolumn{2}{r|}{\it{Single String}}
      & \multicolumn{2}{r|}{\it{Multiple Strings}}
      & \multicolumn{1}{r}{{\it{Average Synonymy}}} \\
      \cline{2-7}
      {\it{Development}} & 1,239 & 1,053 &(85\%)& 186& (15\%) & 2.49 \\
      \cline{1-7}
      {\it{Evaluation}}  &   871 &   853& (98\%)&  18 &(2\%) & 2.06 \\
      \cline{2-7}
     \end{tabular}}
  \end{center}
  \caption{Synonymy of target entities in the development and evaluation datasets, according to the gold standard.}
  \label{tab:synonymy}
\end{table}

\begin{table}[t]
  \begin{center}
    \begin{tabular}{r|r|r|r|r|}
\multicolumn{1}{r}{}
           & \emph{Unique Synsets} & \emph{Single string} & \emph{Multiple Strings} & \multicolumn{1}{r}{\emph{Average Synonymy}}\\ \cline{2-5}
\emph{Nouns}      & 101	    & 30	    & 71	       & 3.34 \\
\emph{Verbs}      & 146	    & 39	    & 107	       & 3.84 \\
\emph{Adjectives} & 31	            & 16	    & 15	       & 3.87 \\ \hline
\emph{Total}      & 278	    & 85	    & 193	       & 3.66 \\ \cline{2-5}
\end{tabular}
\end{center}
\caption{Synonymy for Senseval-3 \emph{lexical sample}, as attested by the gold standard.}
\label{tab:synonymy-gs}
\end{table}

Table~\ref{tab:synonymy-gs} shows average synonymy for a WSD dataset~(Senseval-3 \emph{lexical sample}).
The gold standard contains 278 senses (for 57 target words), of which 85
have a unique lexicalization~(and 193 have multiple), with 3.66 synonyms on average.
Average synonymy in this WSD setting is slightly higher than that of TAC-KBP's gold standard,
similarly to polysemy~(though
\S\ref{sec:ambig-synonymy-dict} will show that, in practice,
synonymy for NED can be much higher,
according to a dictionary ).


To summarize, the number of cases where annotators could not assign
an entity (NILs) is significantly higher in NED than in WSD (around 50\%
compared to just 1.7\%). And ambiguity and synonymy, according to gold
standards, are substantially lower than in WSD~(with average ambiguities
of around 2 {\it{vs}.\ }5, and average polysemies
of 2 {\it{vs.}\ }4). These statistics are
misleading, however, since TAC-KBP's actual ambiguity and synonymy are
more extreme, according to our dictionaries~(discussed in \S\ref{sec:ambig-synonymy-dict}).
Finally, inter-annotator agreement for TAC-KBP is higher --- ranging
between 87\% and 93\%, depending on the entity type --- compared
to 72\% in WSD.

\subsection{Ambiguity and Synonymy in Dictionaries}
\label{sec:ambig-synonymy-dict}

\begin{table}[t]
  \begin{center}
    {\small\begin{tabular}{r|r||r|r||r|r|l}
      \multicolumn{1}{c}{\it{}}
      & \multicolumn{1}{c||}{\it{Unique}}
      & \multicolumn{1}{r|}{\it{No Entities}}
      & \multicolumn{1}{r||}{\it{Single Entity}}
      & \multicolumn{2}{c}{\it{Multiple Entities}} \\
      \cline{3-6}
      \multicolumn{1}{c}{\it{}}
      & \multicolumn{1}{r||}{\it{Strings}}
      & \multicolumn{1}{r|}{{\it{(}}NIL{\it{s)}}}
      & \multicolumn{1}{r||}{\it{(Monosemous)}}
      & \multicolumn{1}{r|}{\it{(Polysemous)}}
      & \multicolumn{1}{r|}{\it{Ambiguity}} \\
      \cline{2-6}
{{LNRM}} & 1,162 & 111 & 186 & 765 & 86.14 \\
      \cline{1-6}
{{HEUR}} & 1,162 & 94 & 331 & 737 & 22.26 \\
      \cline{1-6}
{{LEX + HEUR}} & 1,162 & 193 & 372 & 597 &  15.44\\
      \cline{1-6}
{{SENSE + HEUR}} & 1,162 & 136 & 344 & 682 &  19.01 \\
      \cline{1-6}
    \end{tabular}}
  \end{center}
  \caption{Ambiguity, according to dictionaries, of target strings in the development dataset: top six rows show figures for the \textbf{LNRM} and \textbf{HEUR} dictionaries; last two rows focus on \textbf{HEUR} only, after discarding entities that lack training examples, according to the \textbf{LEX} and \textbf{SENSE} methods~(see \S\protect\ref{sec:superv-disamb}).}
  \label{tab:dict:ambiguity}
\end{table}

We now consider polysemy and synonymy in our dictionaries. Table~\ref{tab:dict:ambiguity} reports
ambiguities for the \textbf{LNRM} and \textbf{HEUR} dictionaries on development
data. For instance, the dev-set comprises 1,162 target strings, of which
\textbf{LNRM} has no suggestions for 111 (94 with \textbf{HEUR}).  Among the remaining
strings, 331 yield a unique suggestion from Wikipedia (331 with \textbf{HEUR}).
The average ambiguity for the 765 strings with multiple candidates
is 86.14 (only 22.26 with \textbf{HEUR}). The distribution of the ambiguity between news and web datasets is distinct, 50.90 and 103.24, respectively.
Thus, \textbf{HEUR} covers more
strings 
and has fewer
monosemous entries 
yet lower polysemy 
than \textbf{LNRM}.  Thus, \textbf{HEUR} retains more references from
Wikipedia and simultaneously substantially reduces ambiguity.  Although we don't tabulate figures for
other dictionaries, we note here that \textbf{FUZZ} has even higher polysemy than \textbf{LNRM}.
The two lower rows in the the table also shows ambiguities faced by our supervised
classifiers, which discard dictionary suggestions for which there
is no training data~(e.g., entities that do not occur as targets of an anchor
link --- see \S\ref{sec:superv-disamb}). The lower half of the table focuses
on the \textbf{HEUR} dictionary in combination with the \textbf{LEX}
and \textbf{SENSE} strategies for gathering training instances.
As expected, resulting ambiguities are lower than for full
dictionaries, with more cases of zero or one candidates
and fewer strings mapping to multiple entities~(also with lower
ambiguities); the decrease
is smaller for the more conservative filtering strategy~(\textbf{SENSE}),
which is again consistent with our expectations.

Ambiguity in dictionaries is much higher than according to the gold
standard~(compare to Table~\ref{tab:ambiguity}).
This comes as no surprise, since the gold standard severely underestimates
true ambiguity by focusing exclusively on entities that are
mentioned in a target dataset. Since many target strings can, in fact,
refer to dozens of possible external entities, our estimates of ambiguity
are also much higher than in WSD settings~(see Table~\ref{fig:senseval3}).  In
contrast to WSD, NED's true ambiguity usually remains unknown, as its
determination would require a laborious inspection of all strings and
entities, and an exhaustive search for examples
of usage in actual text.  Although our dictionaries have good quality
and come close to covering all entities in the gold standard~(see \S\ref{sec:upperb-dict-superv}),
a manual inspection showed that they also contain incorrect entries.
Therefore, we suspect that actual ambiguity maybe be slightly lower
than our estimate obtained with the \textbf{HEUR} dictionary.


\begin{table}[t]
  \begin{center}
    {\small\begin{tabular}{r|r||r|r||r|r|l}
      \multicolumn{1}{c}{\it{}}
      & \multicolumn{1}{r||}{\it{Unique}}
      & \multicolumn{1}{c|}{\it{No}}
      & \multicolumn{1}{c||}{\it{Single}}
      & \multicolumn{1}{c|}{\it{Multiple}}
      & \multicolumn{1}{l}{\it{Average}} \\
      \multicolumn{1}{c}{\it{}}
      & \multicolumn{1}{r||}{\it{Entities}}
      & \multicolumn{1}{c|}{\it{Strings}}
      & \multicolumn{1}{c||}{\it{String}}
      & \multicolumn{1}{c|}{\it{Strings}}
      & \multicolumn{1}{l}{\it{Synonymy}} \\
      \cline{2-6}
       \cline{2-6}
\multicolumn{1}{l|}{EXCT} & 1,239 & 279 (23\%) & 0 & 960 (77\%)& 210.38 \\
      \cline{1-6}
\multicolumn{1}{l|}{EXCT-HEUR} & 1,239 & 296 (24\%) & 0 & 943 (76\%) & 46.37 \\
      \cline{1-6}
    \end{tabular}}
  \end{center}
  \caption{Synonymy of target entities in the development dataset for the EXCT dictionary, with and without filtering heuristics;
since some of these entities are not in Wikipedia, they cannot be suggested by our dictionaries.  } 
  \label{tab:dict:synonymy}
\end{table}

\begin{table}[t]
  \begin{center}
    \begin{tabular}{r|r|r|r|r|}
\multicolumn{1}{r}{} & \emph{Unique Synsets} & \emph{Single String} & \emph{Multiple Strings} &
\multicolumn{1}{r}{\emph{Average Synonymy}}\\ \cline{2-5}
\emph{Nouns}      & 114	    & 34	    & 80	       & 3.35 \\           
\emph{Verbs}      & 202	    & 58	    & 144	       & 3.76 \\           
\emph{Adjectives} & 53	            & 30	    & 23	       & 4.04 \\ \hline
\emph{Total}      & 369	    & 122	    & 247	       & 3.66 \\ \cline{2-5}
\end{tabular}
\end{center}
\caption{Synonymy for Senseval-3 \emph{lexical sample}, as attested by the dictionary.}
\label{tab:synonymy-dict}
\end{table}

Table~\ref{tab:dict:synonymy} shows synonymy figures for both the raw
\textbf{EXCT} dictionary and also after applying heuristic rules~(\textbf{EXCT-HEUR}).
Entities corresponding to NILs were not covered by the dictionary~(tallied under the
\emph{No Strings} heading), and all of the remaining entities
were lexicalized by a large number of strings~(none by just one).
The \textbf{EXCT} dictionary had, on average, 210 strings, which is reduced to 46 with heuristics. 
Although high estimates reflect the
comprehesive coverage afforded by our dictionaries, they do not reveal
true levels of synonymy, which would require hand-checking all entries,
as before.  Nevertheless, these figures illustrate, at a high level,
another important difference between NED and WSD, where synonymy tends to be
much lower~(e.g., around 3.5, according to one popular dictionary
--- see Table~\ref{tab:synonymy-dict}).  Extending the table to include
\textbf{LNRM} and \textbf{FUZZ} dictionaries would require performing
more complicated calculations, but the resulting synonymy estimates
would only be higher for these omitted cascades.

\begin{figure}[t]
\centering
\includegraphics[width=\textwidth]{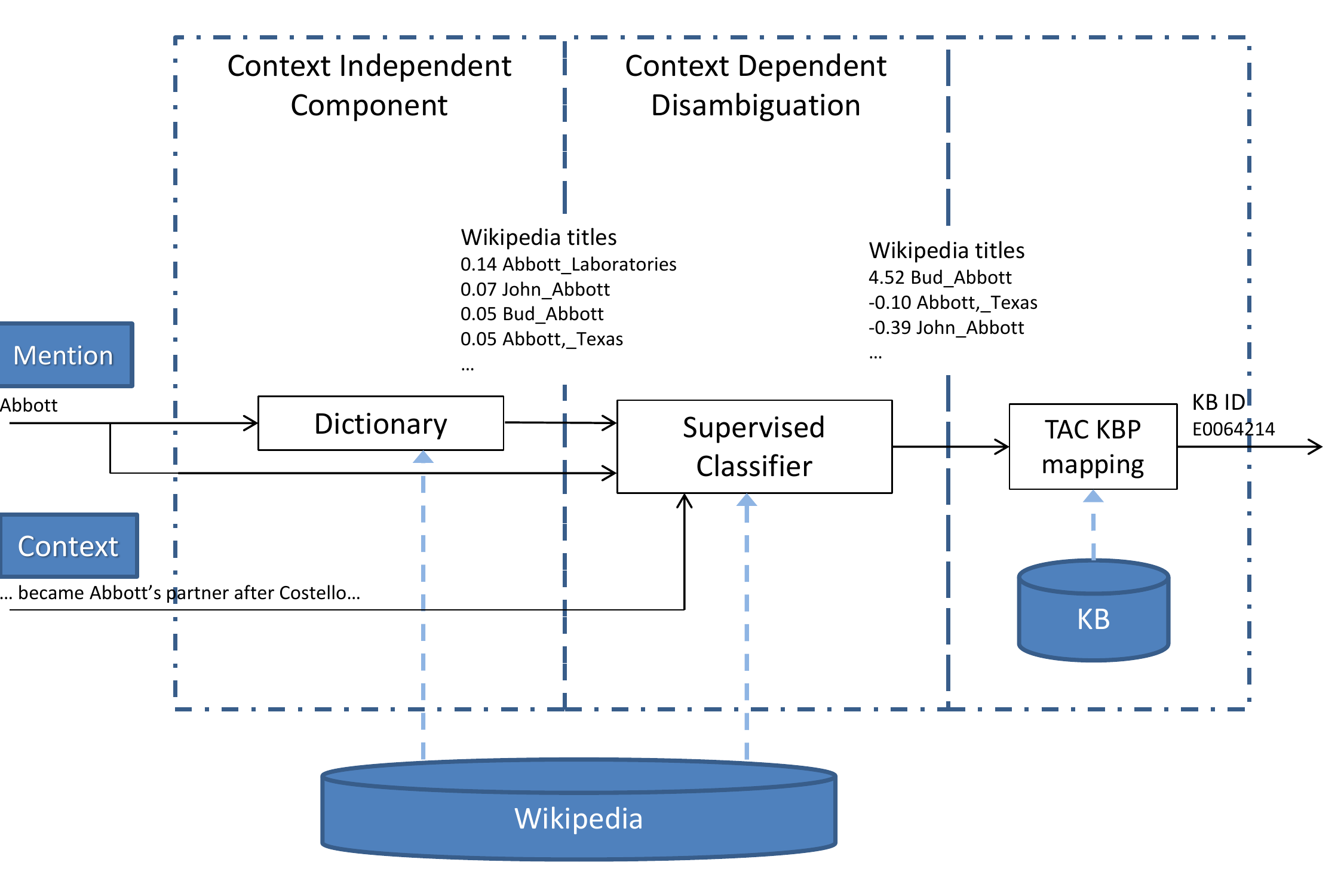}
\caption{Architecture for our system adapted to TAC-KBP.}\label{fig:archtackbp}
\end{figure}

\section{A NED System for TAC-KBP}
\label{sec:ned-system-tac}

Evaluating our system on the TAC-KBP exercise required several
adaptations.  Figure~\ref{fig:archtackbp} shows the updated architecture.
In it, a new module looks up top-ranked articles in the KB.
If a top-scorer has a corresponding entry in the KB, the module returns its KB ID. 
Since all KB-instances came from Wikipedia,
there is a direct mapping from titles to KB IDs.  To guarantee matches,
we altered our dictionary construction process slightly, making sure
to include all Wikipedia titles explicitly referenced by the
official KB~(see Figure~\ref{fig:kbexample}).  When deciding which
pages are canonical~(see \S\ref{sec:redir-canon-articl}), we
preferred entries listed in the KB over all others (from a
superset of Wikipedia data that included the TAC-KBP
dump of October 2008).




\section{Experimental Results and Performance Analyses}
\label{sec:development}

We now evaluate different variants of
dictionaries and supervised classifiers on development
data. These experiments allow us to tune settings, to be
used with the blind test set~(see \S\ref{sec:results}). 

\subsection{Evaluation Setup for TAC-KBP}
\label{sec:evaluation-tac-kbp}

We follow a standard evaluation procedure, using scripts from the 2009--10 TAC-KBP
exercise. The metric is (micro-averaged) accuracy: given a reference
set with $N$ queries --- and a corresponding set of guesses, where
$C$ are correctly disambiguated (i.e., a system's output label equals
the gold standard's entity ID string) --- the score is simply $C/N$.
Since our main goal is to disambiguate among entities,
we will focus on evaluation of the dictionaries and supervised
classifiers restricted to entities present in the
KB~(i.e., ignoring gold standard examples tagged
with NILs).

\subsection{Performance of Top Dictionary Entries}
\label{sec:dictionaries}

\begin{table}[t]
\begin{center}\begin{tabular}{r|r|}
\cline{2-2}
{{EXCT}}  & 0.6937 \\
\hline\hline
{{LNRM}} & 0.6949  \\
\hline
{{FUZZ}}  & 0.7134         \\    
\hline
{{HEUR}}  & 0.7212          \\   
\hline\hline
{{GOOG}}  & 0.6955         \\    
\cline{2-2}         
\end{tabular} \mbox{ }\mbox{ }\mbox{ }\mbox{ }\mbox{ }\mbox{ } \begin{tabular}{r:r}
Wikipedia counts only
& web counts only \\    
\cdashline{1-2}
0.6513     
& 0.6758 \\
\multicolumn{1}{c}{} \\
\multicolumn{1}{c}{} \\
\multicolumn{1}{c}{} \\
\end{tabular}\end{center}
\caption{Performance of dictionaries on the \emph{news} subset of the development dataset --- as micro-averages --- for two individual dictionaries~(\textbf{EXCT} and \textbf{GOOG}), the two dictionary cascades~(\textbf{LNRM} and \textbf{FUZZ}) and our heuristic dictionary~(\textbf{HEUR}).  For \textbf{LNRM}, we show also results with partitions of counts~(i.e., just the popularity within Wikipedia versus on the rest of the web).}
\label{table:dict.results}
\end{table}

Table~\ref{table:dict.results} shows performance on the \emph{news} subset of development data
for the \textbf{EXCT} and \textbf{GOOG} dictionaries~(69.4 and 69.6\%),
the \textbf{LNMR} and \textbf{FUZZ} cascades~(69.5 and 71.3\%),
and the heuristic combination, \textbf{HEUR}~(72.1\%).
It confirms our intuition~(see \S\ref{sec:dictionary}): \textbf{HEUR} improves over the
\textbf{FUZZ} cascade,
offering cleaner suggestions~(since it yields many fewer candidates).
Results from \textbf{GOOG} are competitive with cascades and may be a good
alternative in situations when full dictionaries are unavailable.\footnote{However,
this may require querying the search engine just-in-time,
using Google's API, with consequent limitations.}
The table also shows that using counts from Wikipedia alone
is worse than relying on counts from the rest of the web --- and that
merging the two sets of counts~(see \S\ref{sec:dictionary}) works
best --- for the \textbf{LNRM} cascade; we use merged counts
in all remaining experiments.  Hovering at 70\%, MFS
heuristics have higher accuracies here than even in typical WSD
settings, where they are known to be strong~(e.g.,
around 55\% --- see \S\ref{sec:word-sense-disamb-1}),
which shows that choosing most popular entries makes
for powerful methods in NED as well.

\subsection{Precision/Recall Curves for Dictionaries}
\label{sec:analys-dict}

\begin{figure*}[t]
\begin{center}
\includegraphics[width=\linewidth]{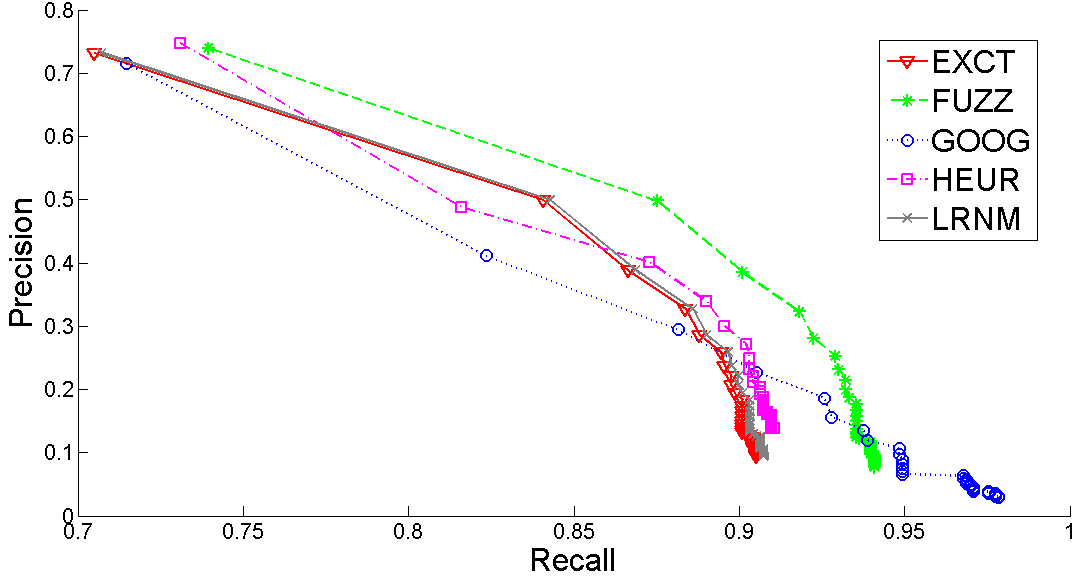}
\caption{Precision and recall @k curves for a random subset of the \emph{news} development dataset (without NILs).}
\label{fig:PaperdevNoNIL.pr}
\end{center}
\end{figure*}

Although performance of top candidates is indicative of a dictionary's
overall quality, it tells us little about the less popular choices.
We use dictionaries to expand queries into pools of possible candidates,
to be disambiguated by a supervised system.  Therefore, it is important
to understand how close a dictionary might come to capturing all --- even
low-ranking --- entities that may be relevant to a gold standard.
Figure \ref{fig:PaperdevNoNIL.pr} shows precision/recall curves
that evaluate our dictionaries beyond just top-ranking
entries~(the left-most points of each curve, corresponding to
Table~\ref{table:dict.results}).\footnote{
Precision is the number of correct
entities divided by the total number of entities returned;
recall is the number of correct
entities divided by the total number
of gold standard entities). The P/R curve is obtained taking the top K candidates with highest probability. }

These curves show that the \textbf{FUZZ}
cascade generates more entities than other dictionaries, with
higher recall at similar levels of precision; unfortunately it
draws on far too many candidates. The \textbf{LNRM} and
\textbf{EXCT} dictionaries perform similarly at all recall levels;
\textbf{HEUR} performs better at high precision, slightly
better at high recall, and worse in-between. The \textbf{GOOG}
dictionary performs worse everywhere except at the highest recall,
where it dominates~(at close to 99\%); its generally-low performance
indicates that although \textbf{GOOG} manages to generate candidates
for nearly all strings, its ranking function is less-well suited
to the NED task than are our methods.


\subsection{Performance of Supervised Classifiers}
\label{sec:superv-class}



Table~\ref{table:test2009.variations} shows performance on the \emph{news} subset of development
data for several variants of our supervised classifier~(described in \S\ref{sec:superv-disamb}).
The first row corresponds to default parameters; the rest represent a greedy exploration of
the space of alternatives.  Each additional row specifies a setting that
differs from the first row; dashes~(\texttt{-}) indicate that all other
parameters are the same.

\begin{table}[t]
\begin{center}\begin{tabular}{|rccccc||r|r|}
\multicolumn{2}{c}{} & \multicolumn{2}{c}{\textbf{Dictionary}} \\
\cline{3-4}
\multicolumn{2}{r}{\textbf{Classifier}\phantom{s}} & \multicolumn{1}{|c}{\emph{Filtering}} & \multicolumn{1}{:c|}{\emph{Back-Off}} & \textbf{Span} & \textbf{Match} & \multicolumn{1}{r}{\emph{news}} \\
\cline{1-7}
\multicolumn{2}{|r}{MAXENT\phantom{s}} & HEUR & LNRM & 100  & LEX   & 0.7707 & \multicolumn{1}{r}{$\Delta$} \\ \hline\hline   
\multicolumn{2}{|r}{multi-class SVM\phantom{s}}
          & -    & -    & -    & -     & 0.7063 & -0.0644 \\          
\multicolumn{2}{|r}{one-versus-all SVMs}
          & -    & -    & -    & -     & 0.7463 & -0.0244 \\ \hline   
\phantom{HACKHA}
&-        & LNRM & -    & -    & -     & 0.7528 & -0.0179 \\          
&-        & -    & HEUR & -    & -     & 0.7827 & +0.0120 \\ \hline   
&-        & -    & -    & SENT & -     & 0.7582 & -0.0125 \\          
&-        & -    & -    & PARA & -     & 0.7582 & -0.0125 \\ \hline   
&-        & -    & -    & -    & SENSE & 0.8090 & +0.0283 \\ \hline   
\end{tabular}\end{center}
\caption{Performance of the supervised classifier on the \emph{news} subset of development data, as micro-averages, for our default supervised classifier~(top row) and several variants, each of which differs from the default in exactly one parameter setting, with the rest indicated by dashes~(\texttt{-}).}
\label{table:test2009.variations}
\end{table}

\noindent
\textbf{Classifier}: The second and third rows correspond to the accuracies of
a multi-class classifier, based on \texttt{SVMmulticlass}~\citep{tsochantaridis04},
and a one-versus-all approach, using \texttt{SVMlight} binary
classifiers~\citep{joachims99}.  Both alternatives perform worse
than our default classification algorithm~(maximum entropy with $\ell_2$-regularization),
MAXENT~\citep{manning-klein03}.  We did not tune any of the available
parameters, since we were interested in out-of-the-box
performance of all methods~(for SVMs, we used a
linear kernel with a cost of 0.01).


\noindent
\textbf{Dictionary}: The fourth row shows that the \textbf{LNRM} cascade
generates worse training examples than our default dictionary combination, \textbf{HEUR}.
We do not show results for other dictionaries, which yield too many candidates~(without
improving precision). The fifth row shows that \textbf{HEUR} also performs better than
\textbf{LNRM} when used as a back-off dictionary, improving over the default.


\noindent
\textbf{Span}: Rows six and seven show that supervised
classification performs equally well using either sentences or
paragraphs~(but that the best results are obtained using left and
right 100 tokens). One reason for similar performance
is that most paragraphs are not marked correctly, in Wikipedia,
often comprising a single sentence.  A fixed span of
tokens to each side of a mention may extend beyond sentence
boundaries, providing more context to help with disambiguation.


\noindent
\textbf{Match}: The last row shows that using all examples that refer
to an entity~(\textbf{SENSE}) improves over the default approach,
which uses only a subset of examples that contain the target string~(\textbf{LEX}).


\subsection{Extending Analyses to Web Data}
\label{sec:news-web}

The best results for the \emph{news} portion of the development data
were obtained with the MAXENT classifier, the \textbf{HEUR} dictionary
for both filtering training examples and backing off, 100-token spans, and
the \textbf{SENSE} match strategy.  On \emph{web} data, our dictionary and
supervised components yield somewhat different results, most of them
congruent with conclusions based on \emph{news} data.

Table~\ref{table:test2009.news.vs.web} shows a subset of the variants
with qualitatively different outcomes.  For instance, the \textbf{LNRM}
dictionary, our default for back-off, is the better option with
\emph{web} data.  As for plain dictionary look-ups, \textbf{HEUR} is still
the best overall choice, although \textbf{GOOG} performs slightly better
on \emph{web} data.  This disparity may stem from search engines being
developed primarily with web-pages in mind, whereas our heuristics came
about from analyzing examples of news articles in the development set;
\textbf{LNRM} may be the more robust back-off dictionary for similar
reasons.
Furthermore, our classifiers fall through to a back-off dictionary
only when there aren't enough training instances, which tends to
be the case for rare entities.  Since \textbf{LNRM} 
has higher recall than \textbf{HEUR}, it may be generally
more useful for obscure references. 
Overall, performance differences between dictionaries are smaller for \emph{web} data, and
average accuracies are substantially higher than with the \emph{news}
portion.  Evaluated against the combined \emph{news}+\emph{web} development
data, ranks of dictionaries remain the same as for \emph{news} alone.

All variations of the supervised system with good performance on
\emph{news} worked better still for \emph{web} data,\footnote{Table~\ref{table:test2009.news.vs.web}
omits results for variations
that did worse than the default on \emph{news}, as they
also do worse with \emph{web} data.} scoring well
above the dictionaries.  The reduction in error going from \emph{news}
to \emph{web} is remarkable, with accuracies well over 90\%.
Manual inspection showed that mentions in \emph{web} data are more
heavily skewed towards most popular entities, compared to \emph{news}
data, which may explain the much higher accuraces of both supervised
classifiers and the underlying dictionaries.  In fact,
\emph{web} data referenced, for the most part, extremely well-known
entities~(e.g., European Union), along with some that are
relatively unknown~(e.g., CCC --- the Cincinnati Cycle Club).
Entities in \emph{news} data, on the other hand,
tended to be of substantially more ambiguous nature.

The second (and final) variant that does better on \emph{news} but did not pan
out with \emph{web} data is the \textbf{SENSE} strategy for gathering
training instances.  Although it has higher coverage~(see \S\ref{sec:ambig-synonymy-dict}),
additional matches tend to be less precise, as these examples aren't required
to contain the target string. Since preliminary results indicate that
relative performance is sensitive to the type of data, we decided to stick with
the simpler, more conservative strategy, \textbf{LEX}. We conclude
that our default settings may already be
optimal for unseen data, and will thus evaluate this set-up against
the TAC-KBP 2010 test data, to compare our
system with the state-of-the-art.

\begin{table}[t]
\begin{center}\begin{tabular}{|ccccc||r|r|}
\multicolumn{5}{r||}{\emph{Dictionaries}}          & \emph{news}   & \multicolumn{1}{r}{\emph{web}}    \\ %
\cline{5-7}
\multicolumn{5}{r||}{{EXCT}}             & 0.6937  & 0.8799   \\   %
\multicolumn{5}{r||}{{LNRM}}             & 0.6949  & 0.8799  \\   %
\multicolumn{5}{r||}{{FUZZ}}             & 0.7134  & 0.8808  \\   %
\multicolumn{5}{r||}{{HEUR}}             & 0.7212  & 0.8845   \\ %
\cline{5-7}
\multicolumn{5}{r||}{{GOOG}}             & 0.6955  & 0.8873  \\   %
\cline{6-7}
\multicolumn{1}{c}{} & \multicolumn{2}{c}{\textbf{Dictionary}} \\
\cline{2-3}
\multicolumn{1}{c}{\textbf{Classifier}} & \multicolumn{1}{|c}{\emph{Filtering}} & \multicolumn{1}{:c|}{\emph{Back-Off}} & \textbf{Span} & \textbf{Match} & \emph{news} & \multicolumn{1}{r}{\emph{web}} \\
\hline
MAXENT   & HEUR & LNRM & 100 & LEX   & 0.7707 & 0.9376 \\ \hline\hline   %
-        & -    & HEUR & -   & -     & 0.7827 & 0.9376 \\ \hline   %
-        & -    & -    & -   & SENSE & 0.8090 & 0.9209 \\ \hline   %
\end{tabular}\end{center}
\caption{Performance on both \emph{web} and \emph{news} subsets of development data, as micro-averages, for dictionaries and three of the supervised classification system variants.} 
\label{table:test2009.news.vs.web}
\end{table}




\subsection{Upper Bounds for Supervised Classification with Our Dictionaries}
\label{sec:upperb-dict-superv}

An important function of the filtering dictionary is to provide sets of plausible
entities, which determine construction of (distantly) supervised training data.
Given an ideal classifier, we would prefer to use comprehensive dictionaries that
might contain the correct entity for all strings in the evaluation
set, even if that meant dragging in many incorrect candidates too.
Table~\ref{table:test2009.oracle} shows the skyline results that
could be attained by an oracle, choosing the best possible entity
available to each system. For \textbf{LNRM} and
\textbf{HEUR} cascades, gold standard entities are among
the dictionaries' suggestions about 92\% of the time~(98\% for \emph{web} data).
Supervised classification with the \textbf{LEX} strategy lowers these bounds
slightly, because in some cases there are no training examples available
for the gold entity, which precludes our classifiers from returning
the correct result; by expanding the pool of training instances,
the \textbf{SENSE} strategy restores this bound to nearly
what it was for \textbf{HEUR}.  Naturally, higher bounds aren't necessarily
superior: for example, \textbf{FUZZ} yields the overall
largest number of possibilities, but has
lower realized performance than \textbf{HEUR}.\linebreak

\begin{table}[t]
\begin{center}\begin{tabular}{ r|rr|rr| }
\multicolumn{1}{r}{}
                & \multicolumn{2}{c|}{\emph{news}} & \multicolumn{2}{c}{\emph{web}}\\            \cline{2-5} %
                         & \emph{realized} & \emph{oracle} & \emph{realized} & \emph{oracle}            \\ \cline{2-5}  %
{{LNRM}} dictionary              & 0.6949 & 0.9158 & 0.8799 & 0.9842   \\   %
{{FUZZ}} dictionary             & 0.7134 & 0.9415 & 0.8808 & 0.9851   \\  %
{{HEUR}} dictionary             & 0.7212 & 0.9188 & 0.8845 & 0.9832  \\  \hline %
SUPERVISED ({{LEX + HEUR}})   & 0.7707 & 0.8955 & 0.9376 & 0.9814  \\   %
SUPERVISED ({{SENSE + HEUR}}) & 0.8090 & 0.9140 & 0.9209 & 0.9832  \\ \cline{2-5}
\end{tabular}\end{center}
\caption{Realized performance and oracle skylines --- on both \emph{news} and \emph{web} portions of development data
--- as micro-averages, for the \textbf{LNRM}, \textbf{FUZZ} and
\textbf{HEUR} dictionaries, as well as for \textbf{LEX} and \textbf{SENSE}
supervised classifier variants~(both also based on \textbf{HEUR}).}
\label{table:test2009.oracle}
\end{table}

\subsection{Final Results and Comparison to the State-of-the-Art}
\label{sec:results}
\label{sec:comparative-results}

\begin{table}[t]
\begin{center}\begin{tabular}{r | r  | r||r|}
\multicolumn{1}{r}{}
                      & \multicolumn{1}{c|}{\emph{news}} 
                      & \multicolumn{1}{c||}{\emph{web}} 
                      & \multicolumn{1}{c}{\emph{full}} \\ \cline{2-4} 
HEUR dictionary         & 0.6984  & 0.8149  & 0.7490  \\          
SUPERVISED (LEX+HEUR)   & 0.8125  & 0.8668  & 0.8448  \\   \cline{1-4}
\end{tabular}\end{center}
\caption{Performance on the 2010 test data, for our best dictionary and supervised classifier, broken down also for the \emph{news} and \emph{web} subsets.}
\label{table:results}
\end{table}

Following the development phase, we tested our best dictionary and supervised classifier
on the hidden (2010 test) dataset.  Table~\ref{table:results} shows both results, also
broken down by data type. 
This final evaluation confirmed
that the heuristic dictionary already performs quite well~(at 75\%) overall,
and that trained classifiers can tap into further improvements~(scoring close to 85\%). The improvement is larger on news, which is again more challenging that the web dataset. The results are similar to those of development, except for web, where results have dropped around 5 points.

\begin{table}[t]
\begin{center}
\begin{tabular}{r|r||l|l}
\multicolumn{1}{c}{} & \emph{System}                    & \emph{KB-only}  \\ 
  \cline{2-4}
 & \citep{dredze-EtAl:2010:PAPERS}   & \phantom{$^{*}$}0.7063  \\ 
 & \citep{Hachey2012}                & $^{*}$0.723             \\ 
 & \citep{varma2009}                 & \phantom{$^{*}$}{0.7654} & Best submission to TAC 2009.\\ 
 & \citep{zhang-EtAl:2010:PAPERS}    & $^{*}$0.792              \\ 
 & \citep{han-sun:2011:ACL-HLT2011}  & $^{*}$0.79               \\ 
 & \citep{guo_graph-based_2011}      & 0.74                    \\
 \cdashline{2-3}
     & HEUR dictionary              & $^{*}$0.7212               \\ 
2009 &  SUPERVISED (LEX+HEUR)       & $^{*}$0.7707               \\ 
  \hline  \hline  
2010 &   \citep{Mcnamee_2010}    &\phantom{$^{*}$}0.6500    & Best ``no context'' system at TAC. \\ 
&   \citep{varma2009}            &\phantom{$^{*}$}0.705     & Reported by \citet{Hachey2012}.\\     
& \citep{guo_graph-based_2011}      & 0.741                    \\
&   \citep{Hachey2012}           &$^{*}$0.784               \\  
&   \citep{citeulike:9118357}    &\phantom{$^{*}$}0.8059    & Best system at TAC 2010. \\ 
\cdashline{2-4}
&   HEUR dictionary              &\phantom{$^{*}$}0.7490         \\
& SUPERVISED  (LEX+HEUR)         &\phantom{$^{*}$}{\bf{0.8448}}  \\
  \cline{2-4}
\end{tabular}
\end{center}
\caption{The state-of-the-art for 2009/2010 TAC-KBP test data, including performance
of our dictionary~(\textbf{HEUR}) and classifier~(\textbf{SUPERVISED}); results from non-blind evaluation
set-ups~(see \S\ref{sec:current-systems}) are starred~($^{*}$).}
\label{tab:comparison-to-literature}
\end{table}

Table~\ref{tab:comparison-to-literature} shows performance of
recent NED systems that report results for either of the 2009/2010
TAC-KBP test sets~(see \S\ref{sec:previous-work} for their descriptions),
including our dictionary and classifier.

The dictionary performs well on the KB-only subset of the 2009 test
set, beating some of the more complex systems.  
Since we
used the 2009 data to choose bests of several system variants,
evaluation on this test set is not blind and may overstate our
results, as well as those of many other published systems that
did the same~\citep{zhang-EtAl:2010:PAPERS,han-sun:2011:ACL-HLT2011,Hachey2012}; 
our supervised classifier scores 77\% on the KB-only subset,
behind two systems~\citep{zhang-EtAl:2010:PAPERS,han-sun:2011:ACL-HLT2011}
scoring 79\%,
both of which were also developed using 2009 test data.

For the 2010 test data, our evaluation was completely blind.  Here
the dictionary performs better than the next best ``no context''
system~\citep{Mcnamee_2010} by a large margin: 75\% {\it{vs}.\ }65\%.  The supervised classifier also 
 scores highest --- 85\% {\it{vs}.\ }81\% for the next best system~\citep{citeulike:9118357}.
The entry for \citet{Hachey2012} represents the best
of several systems chosen on the test set; their best variant according to development data was,
in fact, the system of \citet{varma2009}, which scores 7\% lower\footnote{Note that, since \citet{Hachey2012} reimplemented three
well-known systems~\citep{DBLP:conf/eacl/BunescuP06,DBLP:conf/emnlp/Cucerzan07,varma2009},
our results also compare favorably to the other two.}  

\section{Discussion}
\label{sec:summary-discussion-final}

We have shown that it is possible to construct an effective dictionary for NED,
covering between 92\% and 98\% of the manually annotated string-entity pairs~(for development data).
Although the ambiguity in such dictionaries varies, it tends to be higher than for the gold standard
in NED and also for a typical WSD dataset.  Since a similar phenomenon is also observed with synonymy,
one might expect NED to pose a more difficult problem than WSD; nevertheless, we observed the
opposite effect in practice.  Our popularity-based dictionary heuristic performs even more strongly
than the MFS baseline in WSD~(75\% in our blind NED evaluation, compared to 55\%); supervised system variants
also score much higher~(84\% {\it{vs}.\ }73\%).  Of course, comparing evaluation numbers across
tasks requires extreme caution.  Nonetheless, we suspect that qualitatively large differences
in performance here indicate that NED has larger numbers of training data, compared to WSD.

High ambiguity and synonymy, together with the large volumes of text data,
make NED computationally more demanding than WSD, both in the scope of regular
memory and disk storage capacities, as well as speed and efficient processing requirements.
Our approach in particular could invoke training of potentially millions of classifiers,
requiring significant engineering effort.  But since each classifier can be trained
independently, parallelization is simple and easy.

We found that more comprehensive dictionaries that provide better coverage aren't always
optimal when it comes to training supervised classifiers, since it is also important to
have enough examples for each candidate entity being suggested.  Instead, a
dictionary that contains the most precise and common choices may work better,
as demonstrated by our cascades of dictionaries.  

All in all, our systems fare well, compared to the state-of-the-art.
The dictionary beats all systems not using context in the 2010 TAC-KBP
task by a large margin. And the supervised system, despite its simplicity,
outperforms other systems as well. More
detailed comparisons are difficult, because many recent papers
lack ablative analyses: even when performance of individual
components is reported, interactions with NILs make
proper comparisons challenging.\footnote{E.g.,
\citet{han-sun:2011:ACL-HLT2011} report end-to-end performance
of their popularity and name model~(roughly equivalent to
our dictionary), combined with a NIL detection system, excluding
KB- or NIL-only results. Their score on the 2009 test set
is far below ours~(50\% {\it{vs}.\ }72\%), but it is not possible to identify the
main reason behind this discrepancy.}$^{,}$\footnote{An exception, \citet{Hachey2012}
provide precision and recall of their candidate generation component for KB
queries~(56.3 and 87.8, respectively) on the 2009 test data~(the \emph{news} subset
of our development set), which can be compared to oracle and realized
performance of \textbf{HEUR}~(91.88 and 72.12); they also report an ambiguity of 7.2,
obtained by dividing the number of candidate entities by the total number
of query strings~(our corresponding figure for \textbf{HEUR} would be 9.08,
if calculated in the same fashion, i.e., differently from the numbers
listed in Table~\ref{tab:ambiguity}).}

As mentioned previously, our dictionaries could be expanded by drawing
on additional sources of data.  This fact should not be overlooked,
since \citet{Hachey2012} found that candidate generation accounted
for most of the performance variation in the systems that
they reimplemented; in particular, acronym handling, using
coreference resolution to find longer mentions, led to substantial improvements.
We could augment our dictionaries via various techniques for mining
acronyms~\citep{varma2009,Mcnamee_2010,citeulike:9118357},
metaphones~\citep{varma2009}, as well as
bolded words in first paragraphs and hatnote templates~\citep{Hachey2012}.
Moving beyond the static dictionary model, it is also possible to exploit
dynamic methods for proposing candidates.  Online techniques can make
use of partial matches between tokens in query strings and entities~\citep{varma2009,Mcnamee_2010,citeulike:9118357},
finite-state transducers~\citep{dredze-EtAl:2010:PAPERS},
matches with longer mentions~\citep{citeulike:9118357},
automatic spelling correction~\citep{zheng-EtAl:2010:NAACLHLT},
Wikipedia search engines and the ``Did you mean...'' functionality~\citep{zhang-EtAl:2010:PAPERS}.
Storing as much information as possible in static dictionaries will make
it easier to debug, replicate and share resources.  But some dynamic
lookups could also be batched to enrich a dictionary just-in-time.
A thorough study of the overlap and contribution of various enhanced
candidate generation methods may make for a fruitful research direction.

With respect to disambiguation, our approach closely followed that of
typical supervised WSD systems, which train a classifier for each
target string.  We used anchor texts in
Wikipedia articles to train logistic regressions, showing that this
method is also appropriate and computationally feasible for NED.
Given their popularity in WSD research, we expected straight-forward
classification techniques to be more prevalent in NED, at least as
baselines.  Methods such as ours may also prove to be useful in combination
with other disambiguation systems which tap on different knowledge sources, e.g. those using the hyperlink structure~\citep{Milne:2008,Moro2014}.

In summary, our NED system can be easily replicated because it uses an
offline dictionary.  Its simplicity allows for a clearer understanding
and ablative analyses, compared with systems that rely on dynamic
candidate generation methods.  As a result, it may make a good platform
for testing and incorporating various modular extensions to NED, such as
 NIL classifiers, candidate generators, similarity-based
techniques, global coherence or coreference resolution.
We designed the system to work with the entire space of
Wikipedia articles and strings: no thresholds or other kinds of parameters
were fine-tuned to the test data.  Given the limited number, scope and complexity
of decisions made even in development, we expect our system to be robust.
Naturally, performance could be further optimized to fit a target
corpus or genre, if desired.

\section{Conclusions and Future Work}
\label{sec:concl-future-work}

We presented a system for NED based on word-experts, which is a
well-understood technique from the WSD literature.  Our system
comprises two components: (1)~a context-independent module, based on
frequencies of entities, which returns most popular candidates; and
(2)~a context-sensitive classifier for each named-entity string that
selects entities that are best suited to a mention's surrounding text.
We show that such a word-expert provides results
which are competitive to the state-of-the-art, as measured on the 2009
and 2010 TAC datasets. 

In the future we would like to extend our work
to other datasets, which will provide further points of
comparison to the state of the art. We would also like to include our
classifiers in more complex NED systems, where complementary
information like link structure~\citep{Moro2014}, similarities between
article texts and mention contexts~\citep{Hoffart2012}, as well as
global optimization techniques~\citep{DBLP:conf/acl/RatinovRDA11} could
further improve the results.


We highlighted many connections between WSD and NED.  In WSD settings,
dictionaries are provided, but NED involves constructing possible
mappings from strings to entities --- a step that \citet{Hachey2012}
showed to be key to success, which we also confirmed experimentally.
The resulting dictionaries exhibit very high synonymy and ambiguity
(polysemy) yet still do not cover many occurrences that ought to be
tagged by a NED system, making the task appear more difficult, in
theory, compared to WSD.  But in practice, the opposite seems to be
the case, due to actual mentions being more heavily skewed towards
popular entities than in WSD, a plethora of available training data in
the form of human-entered anchor-texts of hyperlinks on the web, and
higher inter-annotator agreement, which indicates more crisp
differences between possible shades of meanings than in WSD.  As a
result, both popularity-based dictionary lookups (MFS heuristics) and
supervised classifiers, which are traditional WSD architectures,
perform better for NED than for WSD. In the future, we would like to
extend our study using datasets which include all mentions in full
documents~\citep{hoffart-EtAl:2011:EMNLP}.


\section*{Acknowledgements}

We thank Oier Lopez de Lacalle and David Martinez,
for the script to extract features, as well as
Daniel Jurafsky and Eric Yeh, for their
contributions to our earliest participation
in TAC-KBP.

Parts of this work were carried out while Eneko Agirre was
visiting Stanford University, with a grant from the Ministry of Science;
Angel~X. Chang has been supported by a SAP Stanford Graduate Fellowship;
Valentin~I. Spitkovsky has been partially supported by NSF grants
IIS-0811974 and IIS-1216875 and by the
Fannie \& John Hertz Foundation Fellowship.
We gratefully acknowledge
the support of Defense Advanced Research Projects Agency~(DARPA)
Machine Reading Program under Air Force Research Laboratory~(AFRL)
prime contract no.~FA8750-09-C-0181. Any opinions, findings, and
conclusion or recommendations expressed in this material are
those of the author(s) and do not necessarily reflect the
view of the DARPA, AFRL, or the US government. 


\bibliographystyle{plainnat}
\bibliography{main}

\end{document}